\documentclass[%
 aip,
 amsmath,amssymb,
 reprint,%
]{revtex4-1}

\usepackage{graphicx}
\usepackage{dcolumn}
\usepackage{bm}

\usepackage[utf8]{inputenc}
\usepackage[T1]{fontenc}
\usepackage{mathptmx}
\usepackage{etoolbox}
\usepackage{enumitem}
\usepackage{booktabs}
\usepackage{nicefrac}
\usepackage{booktabs}
\usepackage{multirow}
\usepackage{array,graphicx}
\usepackage[dvipsnames, table]{xcolor}
\allowdisplaybreaks
\definecolor{DynCyan}{rgb}{0.1058, 0.4549, 0.5686}
\makeatletter
\newcommand{\setword}[2]{%
  \phantomsection
  #1\def\@currentlabel{\unexpanded{#1}}\label{#2}%
}

\makeatletter
\def\@email#1#2{%
 \endgroup
 \patchcmd{\titleblock@produce}
  {\frontmatter@RRAPformat}
  {\frontmatter@RRAPformat{\produce@RRAP{*#1\href{mailto:#2}{#2}}}\frontmatter@RRAPformat}
  {}{}
}%
\makeatother
\allowdisplaybreaks
\begin{document}

\preprint{AIP/123-QED}

\title[]{Bridging ocean wave physics and deep learning: Physics-informed neural operators for nonlinear wavefield reconstruction in real-time}
\author{Svenja Ehlers}
\email{svenja.ehlers@tuhh.de}
\affiliation{ 
Dynamics Group, Hamburg University of Technology, Germany
}%
\author{Merten Stender}%
\affiliation{ Cyber-Physical Systems in Mechanical Engineering, Technische Universität Berlin, Germany}%

\author{Norbert Hoffmann}
\affiliation{ 
Dynamics Group, Hamburg University of Technology, Germany
}
\affiliation{ Department of Mechanical Engineering, Imperial College London, United Kingdom}%

\date{\today}

\begin{abstract}
Accurate real-time prediction of phase-resolved ocean wave fields remains a critical yet largely unsolved problem, primarily due to the absence of practical data assimilation methods for reconstructing initial conditions from sparse or indirect wave measurements. While recent advances in supervised deep learning have shown potential for this purpose, they require large labelled datasets of ground truth wave data, which are infeasible to obtain in real-world scenarios. To overcome this limitation, we propose a Physics-Informed Neural Operator (PINO) framework for reconstructing spatially and temporally phase-resolved, nonlinear ocean wave fields from sparse measurements, without the need for ground truth data during training. This is achieved by embedding residuals of the free surface boundary conditions of ocean gravity waves into the loss function of the PINO, constraining the solution space in a soft manner. After training, we validate our approach using highly realistic synthetic wave data and demonstrate the accurate reconstruction of nonlinear wave fields from both buoy time series and radar snapshots. Our results indicate that PINOs enable accurate, real-time reconstruction and generalize robustly across a wide range of wave conditions, thereby paving the way for operational, data-driven wave reconstruction and prediction in realistic marine environments.
\end{abstract}

\maketitle

\section{Introduction}

Real-time and phase-resolved wave prediction of nonlinear ocean surface waves has the potential to enhance the planning of offshore operations, optimizing wave energy converter control, and improve maritime safety. In this context, \textit{phase-resolved} refers to the precise tracing of the wave evolution in space and time. \textit{Real-time capability} requires wave computations to be faster than the physical waves, while modelling \textit{nonlinearity} is crucial to capture key dynamics such as wave-wave interactions, nonlinear dispersion, and rogue waves. 
An important aspect of phase-resolved wave prediction is that, in practice, the initial conditions are typically not available. Wave fields are measured using in-situ (e.g. wave rider buoys) or remote sensing (e.g. radar systems) instruments, resulting in sparse measurement data. This first requires a reconstruction of the corresponding wave field from these measurements to initiate the actual wave prediction.

For intermediate to deep water, the nonlinear potential flow theory is commonly used to model surface gravity waves and can be solved with various numerical methods \cite[cf.][]{Engsig2009, Wu1994, Ma2006, Grilli2001}. 
The High-Order Spectral Method (HOSM) \cite{Dommermuth1987, West1987} is popular for this purpose \cite[cf.][]{Wu2004, Ducrozet2007, Ducrozet2016, Bonnefoy2010}, as with ongoing advancements in computational hardware, HOSM holds promise for highly accurate real-time predictions of nonlinear water waves \cite{Klein2020, Ducrozet2012, Koellisch2018}. 
However, the fast and accurate reconstruction of the initial conditions, also known as wave data assimilation, remains a more difficult task \cite{ Koellisch2018, Hlophe2023}: HOSM-assimilation methods follow either sequential \cite{Yoon2016, Wang2021} or variational \cite{Blondel2010, Aragh2008, Qi2018, Fujimoto2020, Wu2022} strategies, which may cause poor accuracy or require computationally expensive cost-function minimization.
Moreover, these methods usually determine only the optimal initial surface elevation, whereas a full potential flow description requires both elevation and surface velocity potential \cite{Desmars2021}. To address this, Desmars et al. (2023) \cite{Desmars2023} and Köllisch et al. (2018) \cite{Koellisch2018} integrate the free surface boundary conditions into their assimilation methods. Although the first method enables effective parallelization, its real-time capability is not demonstrated. While the second approach theoretically allows for real-time computations, it instead demands unrealistically high spatial and temporal resolution of measurement data \cite{Desmars2020}.

In summary, the data assimilation problem for initializing nonlinear, phase-resolved, real-time wave prediction methods remains largely unresolved. Inspired by recent advances in artificial intelligence that have transformed various research domains, our previous work explored (physics-informed) machine learning approaches to address this challenge: 
\begin{itemize}[leftmargin=*]
\setlength{\itemsep}{0pt}
    \item In Ehlers et al. (2023) \cite{Ehlers2023}, we demonstrate that a Fourier Neural Operator (FNO) architecture \cite{Li2020} effectively reconstructs initial wave elevations from radar snapshots. Once trained, the FNO generalizes this mapping across diverse wave conditions and enables real-time applications. However, its supervised training approach requires a large dataset of ground truth wave elevations for training. While radar measurements are available in real-world settings, acquiring high-resolution ground truth wave elevation data would necessitate an impractically dense network of buoys. Therefore, the FNO-approach relies on synthetic wave and radar data and cannot be retrained using real-world data.
    \item In Ehlers et al. (2025) \cite{Ehlers2025}, we address the unavailability of ground truth wave data by utilizing a Physics-Informed Neural Network (PINN) \cite{Raissi2019}. Unlike the FNO-approach, the PINN infers high-resolution spatio-temporal wave elevations and velocity potentials directly from sparse measurement data, without access to a ground truth. The physical consistency of these reconstructed wave surfaces is ensured by incorporating the potential flow equations into the loss function. However, PINNs require direct measurements of the physical quantities defining the solution. This implies, PINNs can assimilate wave surfaces from buoy measurements, but not from indirect observations such as radar backscatter intensities. Moreover, PINNs lack generalization across wave conditions and solve an individual inverse problem for each wave instance, causing high computational costs that limit their applicability in real-time scenarios.
\end{itemize}
Also, ML-based approaches for modelling phase-resolved wave dynamics proposed by other research groups \cite{Law2020, Zhang2022, LeQuang2023, Liu2024} have also not yet resolved these challenges: Trained using only one or a few specific sea states, they are prone to poor generalization. Moreover, these methods map wave time-series to time-series at fixed spatial points, preventing extrapolation of wave information to untrained locations. This implies the methods can be considered temporally phase-resolved but not spatially. While more advanced ML methods \cite{Mohaghegh2021, Wedler2023} can represent spatio-temporal wave dynamics, they assume fully resolved initial conditions, which is impractical for real-world scenarios.

Therefore, we aim to advance the reconstruction of initial wave fields from sparse measurements by combining the strengths of our FNO-based \cite{Ehlers2023} and PINN-based \cite{Ehlers2025} approach, leveraging the concept of Physics-Informed Neural Operators (PINOs) proposed by Li et al. (2024) \cite{Li2024}. PINOs are a variation of neural operators that incorporate physical laws into their loss functions to overcome the limitations of purely physics-based
and purely data-driven methods. The approach has been validated on PDE benchmarks \cite{Rosofsky2023} and applied e.g. to fluid dynamics \cite{Zhao2024}, acoustics \cite{Konuk2021}, or chaotic system dynamics \cite{Wang2024}. With the PINO acting as an universal approximator for a nonlinear solution operator \cite{Chen1995}, we hypothesize that the PINO-based wave reconstruction approach developed in this work 
\begin{itemize}
\setlength{\itemsep}{0pt}
    \item [(\setword{H1}{hyp:1})] reconstructs spatially and temporally phase-resolved, nonlinear wave surfaces from (a) sparse buoy time-series or (b) radar snapshot measurements 
    \item [(\setword{H2}{hyp:2})] eliminates the need for ground truth data during training, enabling learning from real-world data
    \item [(\setword{H3}{hyp:3})] allows for rapid wave surface reconstruction, supporting real-time forecasting applications
    \item [(\setword{H4}{hyp:4})] generalizes its reconstruction ability for a broad range of wave conditions
\end{itemize}
To validate these hypotheses, Sec. \ref{sec:methods} introduces the high-order spectral method used to generate realistic wave reference data, followed by wave measurement techniques and a description of the PINO architecture and training procedure. The PINO approach is evaluated in Sec. \ref{ch:results_buoy}  for reconstructing wave surface elevations from buoy time-series measurements, and in Sec \ref{ch:results_radar} for reconstruction from radar intensity snapshots, using only a single time-series measurement for calibration.  Finally, Sec. \ref{ch:conclusion} summarizes the results and discusses the potential for operational deployment of the PINO approach.

\section{Methods}
\label{sec:methods}

This section first introduces the High-order spectral method used to generate synthetic wave elevation and surface potential data that are used as reference only to compare the PINO reconstructions post-training. Next, we describe how we generate sparse buoy or radar measurements intended as the input. The PINO architecture tailored to the assimilation problem and its training procedure are explained in the last subsection.

\subsection{Potential flow theory and high-order spectral  method}
\label{ch:HOSM}

In the modelling of nonlinear, non-breaking ocean gravity waves, the problem is typically formulated within the framework of potential flow theory. Assuming an inviscid and irrotational fluid, the flow field is described by a velocity potential $\Phi(x,z,t)$ in the fluid volume and a surface elevation $\eta(x,t)$, governed by the Laplace equation, a kinematic and dynamic surface boundary condition and the bottom boundary at depth $d$.
Solving this system of equations analytically is challenging, such that numerical methods are used to approximate solutions. The High-Order Spectral Method (HOSM) \cite{Dommermuth1987, West1987} has proven particularly efficient and accurate for that purpose \cite{Klein2020, Chiang2005}.

The HOSM first introduces a surface velocity potential $\Phi^\mathrm{s}(x,t) = \Phi (x, \eta(x,t), t)$, which allows us to convert the potential flow problem from the fluid domain to free surface quantities by rewriting it in the form of Zakharov \cite{Zakharov1968} 
\begin{small}\begin{align}
    \eta_t+ \eta_x\Phi^\mathrm{s}_x-W\left( 1+\left(\eta_x\right)^2\right) &=0 \hspace{0.4cm}      \text{on} \;\; z=\eta(x, t)  \label{eq:kinFSBC}\\ 
    \Phi^\mathrm{s}_t +g \eta +\frac{1}{2}\left(\Phi^\mathrm{s}_x\right)^2 - \frac{1}{2}W^2\left( 1+\left(\eta_x\right)^2\right) &=0 \hspace{0.4cm}      \text{on} \;\; z=\eta(x, t)  \label{eq:dynFSBC}.
\end{align}\end{small}
 For these free surface boundary conditions (FSBC) and initial values of $\eta$ and $\Phi^\mathrm{s}$, numerical integration in time can be performed, with the vertical velocity on the free surface $W=\Phi_z|_{z=\eta(x,t)}$ being the only unknown remaining. To solve for $W$, the HOSM expresses the velocity potential and vertical velocity as perturbation series in wave steepness
\begin{small}\begin{align} 
     \Phi(x,z,t) &= \sum_{m=1}^{M} \Phi^{(m)}(x,z,t) \label{eq:PerturbationPhi}\\
     W(x,t) &= \sum_{m=1}^{M} W^{(m)}(x,t)\label{eq:PerturbationW}
\end{align}\end{small}up to a predefined order of nonlinearity $M$. This allows expansion of each $\Phi^{(m)}$ and $W^{(m)}$ as a Taylor series around the mean free surface. Separating the terms of each order $(m)$ yields
\begin{small}\begin{align}
    \Phi^{(1)}(x,0, t)&=\Phi^\mathrm{s}(x,t) \label{eq:TaylorPhiS}\\    
    \Phi^{(m)}(x,0,t)&= -\sum_{\ell=1}^{m-1} \left. \frac{\eta^\ell}{\ell!} \frac{\partial^\ell \Phi^{(m-\ell)}}{\partial z^{\ell}} \right|_{z=0}
\end{align}\end{small}for the perturbation potentials. The terms of the vertical velocity are obtained by a further derivative in $z$-direction
\begin{small}\begin{align}
    W^{(m)}(x,t) = \sum_{\ell=0}^{m-1} \left. \frac{\eta^{\ell}}{\ell!} \frac{\partial^{\ell+1} \Phi^{(m-\ell)}}{\partial z^{\ell+1}} \right|_{z=0} \label{eq:TalorW}.
\end{align}\end{small}
Assuming periodic boundaries, the Fast Fourier Transform can be used to determine $\Phi^{(m)}$ and its spatial derivatives in Fourier space, while the nonlinear products are calculated in physical space. In most engineering applications, $M=4$ is sufficient to capture all important nonlinear interactions \cite{Desmars2020, Luenser2022}.

\subsection{Wave Measurements}
\label{ch:wave_measurments}

Wave measurement systems can be broadly categorized into in-situ (e.g. wave rider buoys, acoustic and pressure sensors) and remote sensing (e.g. X-band radar, Lidar cameras) methods. This work focuses on buoy (case A) and X-Band radar measurements (case B) as input for wave reconstruction. 

In-situ devices provide single-point measurements, which lack directional wavefield information, often necessitating an array of several devices \cite{Klein2020, Hlophe2023}. Instead, when desiring high spatial resolution, measurements are usually performed using remote sensors mounted on a structure.
The antenna of an X-Band radar system rotates with a revolution time of usually $1-2\, \mathrm{s}$ \cite{Neill2018} while emitting radar beams that interact with the ocean surface, causing a backscatter to the antenna \cite{Valenzuela1978}. This way, several square kilometres are monitored with a spatial resolution of approximately $4-10\, \mathrm{m}$  \cite{Neill2018}. However, radar backscatter intensities $\xi$  do not directly correlate with wave elevations $\eta$ due to modulation mechanisms: Tilt modulation depends on the local incidence angle, which affects the radar's ability to detect backscatter based on the relative wave facet orientation. Shadowing modulation occurs when high waves prevent the radar beams from reaching waves behind, leading to zero-backscatter intensities \cite{Dankert2004, NietoBorge2004, Salcedo-Sanz2015}. 

\begin{figure}[ht]

\centering
\includegraphics[scale=0.9]{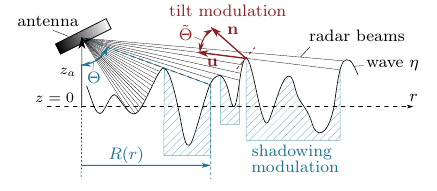}
\vspace{-0.2cm}
\caption{Geometric representation of tilt and shadowing modulation.} 
\label{fig:tilt+shadowing}
\end{figure}

Since real-world measurements lack ground truth wave elevation data for direct comparison, this study validates the PINO approach using synthetic data. To generate synthetic X-Band radar data from wave elevation surfaces, we use a geometric approach depicted in Fig. \ref{fig:tilt+shadowing}. Tilt modulation $\mathcal{T}$ is modelled along a range $r$ around the radar antenna and based on the local incidence angle $\Tilde{\Theta}(r,t)$, defined between the normal vector $\mathbf{n}(r,t)$ of the illuminated wave facet $\eta(r,t)$ and the vector $\mathbf{u}(r,t)$ pointing toward the antenna as  
\begin{small}\begin{equation}
    \mathcal{T}(r,t) =
    c_1 \, \frac{\mathbf{n}(r,t) \, \mathbf{u}(r,t)}{|\mathbf{n}(r,t)| \, |\mathbf{u}(r,t)|} +c_2 \hspace{0.5cm}   \text{if} \;\; |\Tilde{\Theta}(r,t)| \le \frac{\pi}{2}
    \label{eq:tilt_modulation}
\end{equation}\end{small}
whereas $|\cdot|$ is the Euclidean norm and $c_1$ and $c_2$ are scaling factors and offsets depending on the system calibration \cite{Desmars2023}.

Shadowing modulation depends on the nominal incidence angle $\Theta(r,t) = \tan^{-1} \left[ \frac{R(r)}{z_\mathrm{a}-\eta(r,t)} \right]$, where $R(r)$ is the horizontal distance from the radar, and $z_\mathrm{a}$ is the antenna height. A wave $\eta(r,t)$ is shadowed if a closer wave $\eta(r',t)$ at $ R(r') < R(r)$ satisfies $ \Theta(r',t) \ge \Theta(r,t)$. The total radar backscatter intensity 
\begin{small}\begin{equation}
\xi(r,t)=
    \begin{cases}
        0   & \hspace{0.5cm} \text{if} \;\; R(r')<R(r) \text{ and } \Theta(r',t) \ge \Theta(r,t) \\
        \mathcal{T}(r,t)   & \hspace{0.5cm} \text{otherwise}
    \end{cases}
    \label{eq:shadowing-mask}
\end{equation}\end{small}
then is a combination of tilt and shadowing effects \cite{NietoBorge2004, Salcedo-Sanz2015}.

\subsection{Physics-informed neural operator}
\label{ch:PINO}

To address the limitations of our previous ML approaches for reconstructing initial wave fields from sparse measurements, we develop a Physics-Informed Neural Operator (PINO). Unlike a purely data-driven approach, the PINO training does not rely on minimizing the error between the network's reconstructions $\tilde{\eta}\in \mathbb{R}^{n_x \times n_t}$ and a high-resolution ground truth $\eta_\mathrm{true}\in \mathbb{R}^{n_x \times n_t}$, as obtaining such spatio-temporal wave data as the ground truth is often impractical in real-world scenarios. Instead, the PINO leverages sparse measurements where available and accounts for the remaining domain by incorporating physics into its loss function.

\subsubsection{Sensor data generation}
\label{sec:data generation}

While our PINO approach eliminates the need for fully resolved ground truth wave surfaces during training, such data remains valuable to validate the physical correctness of the reconstructions after training. Once validated, the PINO could be trained directly on real-world measurements. 
To enable this validation, we generate synthetic wave elevation surfaces $\eta_\mathrm{HOSM}$ and corresponding surface potentials $\Phi_\mathrm{HOSM}$ using the HOSM described in Sec. \ref{ch:HOSM}. The initial elevations are sampled from a JONSWAP TMA spectrum \cite{Hasselmann1973, Bouws1985} with random phases, a peak enhancement factor $\gamma=3.3$ and various combinations of peak wavelength $L_\mathrm{p}\in \{100, \,110, \hdots, 200\} \,\mathrm{m}$ and wave steepness $\epsilon \in \{0.02, 0.03, \hdots, 0.13\}$.
A relaxation period of 20 peak periods precedes data collection to allow the development of nonlinear wave dynamics. The saved domain for $\eta_\mathrm{HOSM}$ and $\Phi_\mathrm{HOSM}$ spans $1953 \,\mathrm{m}$ with resolution $\Delta x=3.906 \, \mathrm{m}$ ($n_x =500$) and $100 \, \mathrm{s}$ with resolution $\Delta t=0.2\, \mathrm{s}$ ($n_t=500$), although the HOSM itself uses smaller time steps.

In total, 1056 wave surfaces $\eta_\mathrm{HOSM} \in \mathbb{R}^{n_x \times n_t}$ and velocity potentials $\Phi^\mathrm{s}_\mathrm{HOSM} \in \mathbb{R}^{n_x \times n_t}$ are generated (eight samples per $\epsilon$-$L_\mathrm{p}$-combination). From these surfaces, $n_\mathrm{buoy}=5$ buoy measurements $\eta_\mathrm{m}\in \mathbb{R}^{n_\mathrm{buoy} \times n_t}$ are extracted as PINO input for Sec. \ref{ch:results_buoy} (case A). For Sec. \ref{ch:results_radar} (case B) instead, radar snapshots are generated as input using Eqs. (\ref{eq:tilt_modulation}-\ref{eq:shadowing-mask}) with a radar rotation period of $\Delta t_\mathrm{r}=2.0 \, \mathrm{s}$. Each sample then contains $n_\mathrm{snap}=50$ snapshots, is denoted as $\xi_\mathrm{m} \in \mathbb{R}^{n_x \times n_\mathrm{snap}}$, and is accompanied by a single buoy measurement $\eta_\mathrm{cal} \in \mathbb{R}^{1 \times n_t}$, which serves for calibration but is not included as a direct input.
The complete dataset is split into training, validation, and test sets at a ratio of $0.6-0.2-0.2$. resulting in a relatively small number of 634 measurement samples for training. Although the PINO is solely trained on sparse measurements $\eta_\mathrm{m}$ or $\xi_\mathrm{m}$, the original, fully-resolved HOSM data is split using the same ratio, enabling a comparison of the PINO’s reconstruction ($\tilde{\eta}$ and $\tilde{\Phi}^\mathrm{s}$) with a reference ($\eta_\mathrm{HOSM}$ and $\Phi^\mathrm{s}_\mathrm{HOSM}$) after training.

\subsubsection{PINO architecture and training}
\label{sec:loss and training}
\begin{figure*}[ht]
\centering
\includegraphics[scale=0.92]{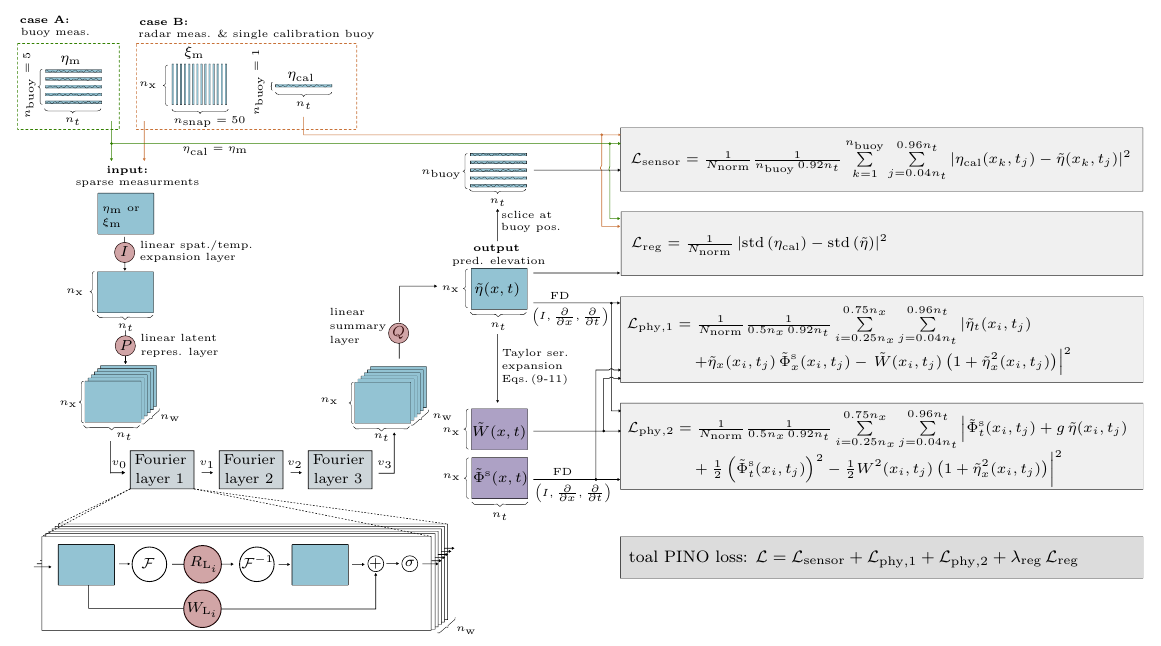}
\caption{Physics-informed neural operator (PINO) for reconstructing spatio-temporal wave elevation fields $\tilde{\eta}$ from sparse buoy $\eta_\mathrm{m}$ (case A) or radar $\xi_\mathrm{m}$ (case B) measurements. Each sparse input sample is lifted to a latent space by layers $I$ and $P$, processed by $n_\mathrm{F}$ Fourier layers and projected to the target output dimension by layer $Q$. Due to the absence of ground truth data,  training is guided by sensor consistency at the calibration buoy locations from the input ($\mathcal{L}_\mathrm{sensor}$), physics-based losses from residuals of the free-surface boundary conditions ($\mathcal{L}_\mathrm{phy,1}$, $\mathcal{L}_\mathrm{phy,2}$), and a regularization term $\mathcal{L}_\mathrm{reg}$ promoting meaningful wave elevation in $\tilde{\eta}$. The required surface potential $\tilde{\Phi}^\mathrm{s}$ and vertical velocity $\tilde{W}$ for the physics-loss terms are approximated from $\tilde{\eta}$ via HOSM (Eq. \ref{eq:PerturbationPhi}-\ref{eq:TalorW}), with spatial and temporal derivatives computed in Fourier space.} 
\label{fig:PINO}
\end{figure*}

 Our PINO architecture in Fig. \ref{fig:PINO} builds upon the strengths of the FNO \cite{Li2020},  as learning features in Fourier space has proven effective for wave reconstruction \cite{Ehlers2023}.
 The model either takes sparse buoy measurements $\eta_\mathrm{m} \in \mathbb{R}^{n_\mathrm{buoy} \times n_t}$ (case A) or radar snapshots $\xi_\mathrm{m} \in \mathbb{R}^{n_x \times n_\mathrm{snap}}$ (case B) as input and outputs high-resolution spatio-temporal wave elevations $\tilde{\eta} \in \mathbb{R}^{n_x \times n_t}$. Unlike standard FNOs, which have matching input and output dimensions, our PINO setup first requires dimension expansion via a linear layer $I$ in space (case A) or time (case B). Another linear layer $P$ then projects this expanded input into a latent representation $v_0 \in \mathbb{R}^{n_x \times n_t \times n_\mathrm{w}}$ with $n_\mathrm{w}$ channels. To reduce boundary effects during the Fourier transform, $v_0$ is padded with 12 zero entries in both dimensions before entering the first Fourier layer.

Each Fourier layer operates on its input $v_i $ via two complementary paths. The upper path applies a discrete Fourier transform $\mathcal{F}$ to each channel, truncates to $n_\mathrm{m}$ modes, scales via a linear transformation $R_{\mathrm{L}i}$ and then transforms back to the spatial domain via inverse Fourier transform $\mathcal{F}^{-1}$. The lower path applies a linear transformation $W_{\mathrm{L}i}$ in the spatial domain to account for non-periodic boundaries and higher-order modes. The outputs of both paths are summed, passed through a GeLU activation $\sigma$, resulting in $v_{i+1} $ for the next layer. After $n_\mathrm{F}$ Fourier layers,  the zero-padding is removed and a final linear layer $Q$ maps the latent representation to the output $\tilde{\eta} \in \mathbb{R}^{n_x \times n_t}$.

To generate meaningful wave surfaces $\tilde{\eta}\in \mathbb{R}^{n_x \times n_t}$ from measurements, the weights of the PINO layers ($I$, $P$, $R_{\mathrm{L}_i}$, $W_{\mathrm{L}_i}$, $Q$) are optimized by minimizing a loss function. In the absence of spatio-temporal ground truth data, alternative loss sources are required.
One source is the sensor loss term $\mathcal{L}_\mathrm{sensor}$, which compares the output $\tilde{\eta} $ to available buoy data. For case A, where buoy measurements $\eta_\mathrm{m}\in \mathbb{R}^{5 \times n_t}$ serve as input, the reconstructed values $\tilde{\eta}_\mathrm{b} \in \mathbb{R}^{5 \times n_t}$ are extracted from $\tilde{\eta}$ at buoy locations $x_\mathrm{b}$ and compared to the input $\eta_\mathrm{cal} =\eta_\mathrm{m}$. In case B, using radar snapshots as input, a single time series $\tilde{\eta}_\mathrm{b} \in \mathbb{R}^{1 \times n_t}$ is extracted at the calibration buoy location and compared to $\eta_\mathrm{cal} \in \mathbb{R}^{1 \times n_t}$.
Next, to achieve physical consistency also between the buoy locations, we incorporate the residuals of the FSBCs (Eqs. \ref{eq:kinFSBC}-\ref{eq:dynFSBC}) as physics-informed loss terms $\mathcal{L}_\mathrm{phy,1}$ and $\mathcal{L}_\mathrm{phy,2}$. These equations involve the surface velocity potential $\tilde{\Phi}^\mathrm{s}\in \mathbb{R}^{n_x \times n_t}$ and vertical velocity $\tilde{W}\in \mathbb{R}^{n_x \times n_t}$ which are approximated from the predicted surface elevation $\tilde{\eta}$ using a fourth-order Taylor series expansion from the HOSM (Eqs. \ref{eq:PerturbationPhi}–\ref{eq:TalorW}). The required spatial and temporal derivatives of $\tilde{\eta}$  and $\tilde{\Phi}^\mathrm{s}$ are computed effectively using differentiation in the Fourier space (FD) \cite{Li2020}. As the HOSM and FD inherently assume periodic boundaries - rarely present in real-world data - we apply Tukey windows along both axes when computing $\tilde{\Phi}^{\mathrm{s}}$, $\tilde{W}$ and derivatives to suppress spectral leakage. Therefore, loss terms are evaluated only over the central region ($i=\nicefrac{n_x}{4}, \hdots,\nicefrac{3 n_x}{4}$ and $j=\nicefrac{n_t}{25}, \hdots,\nicefrac{24 n_t}{25}$), while larger data structures are processed in the background.
Finally, to avoid trivial solutions (e.g. $\tilde{\eta}=0\, \mathrm{m}$, which satisfies the physics losses), we introduce a regularization term $\mathcal{L}_\mathrm{reg}$ that compares the standard deviation ($\mathrm{std}$) of the output field $\tilde{\eta}$ to that of the sparse calibration buoy $\eta_\mathrm{cal}$, both only inside the region unaffected by the Tukey window. While not strictly necessary and weighted modestly by $\lambda_\mathrm{reg}=0.25$, the term $\mathcal{L}_\mathrm{reg}$ promotes realistic variability in buoy-free regions, enhancing accuracy and convergence. In summary, the total loss
\begin{small}\begin{equation}
\mathcal{L} = \mathcal{L}_\mathrm{sensor} + \mathcal{L}_\mathrm{phy,1} + \mathcal{L}_\mathrm{phy2} + \lambda_\mathrm{reg}\, \mathcal{L}_\mathrm{reg} \label{eq:L_total}
\end{equation}\end{small}
then is composed of multiple components \begin{small}\begin{widetext}\begin{align}
    \mathcal{L}_\mathrm{sensor} &= \tfrac{1}{N_\mathrm{norm}} \,\tfrac{1}{n_\mathrm{buoy} \, n_t} \sum_{k=1}^{n_\mathrm{buoy}} \sum_{j=0.04n_t}^{0.96n_t} \left|\eta_\mathrm{cal} \left(x_k, t_j \right) -\tilde{\eta}_\mathrm{b}\left(x_k, t_j \right) \right|^2 \label{eq:L_sensor} \\
    \mathcal{L}_\mathrm{phy,1} & = \tfrac{1}{N_\mathrm{norm}} \, \tfrac{1}{0.5n_x \, n_t} \sum_{i=0.25 n_x}^{0.75 n_x} \sum_{j=0.04n_t}^{0.96n_t} \left| \tilde{\eta}_t(x_i, t_j) + \tilde{\eta}_x(x_i, t_j) \, \tilde{\Phi}^\mathrm{s}(x_i,t_j) -\tilde{W}(x_i, t_j)\left(1+\tilde{\eta}_x^2(x_i, t_j) \right)  \right|^2 \label{eq:L_p1}\\
    \mathcal{L}_\mathrm{phy,2} & = \tfrac{1}{N_\mathrm{norm}} \, \tfrac{1}{0.5n_x \, n_t} \sum_{i=0.25 n_x}^{0.75 n_x} \sum_{j=0.04n_t}^{0.96n_t} \left| \tilde{\Phi}^\mathrm{s}_t(x_i, t_j) +g \tilde{\eta}(x_i,t_j) + \tfrac{1}{2}\left(\tilde{\Phi}^\mathrm{s}_t (x_i, t_j) \right)^2+\tfrac{1}{2}\tilde{W}^2(x_i, t_j)\left(1+\tilde{\eta}_x^2(x_i, t_j) \right)  \right|^2 \label{eq:L_p2}\\
    \mathcal{L}_\mathrm{reg} & = \tfrac{1}{N_\mathrm{norm} } \left| \mathrm{std}(\eta_\mathrm{cal}) -\mathrm{std}(\tilde{\eta}) \right|^2. \label{eq:L_reg}
\end{align}
\end{widetext}
\end{small}The sample-specific normalization factor
\begin{small}\begin{equation}
    N_\mathrm{norm} = \tfrac{1}{n_\mathrm{buoy} \, n_t} \sum_{k=1}^{n_\mathrm{buoy}} \sum_{j=0.04 n_t}^{0.96n_t} |\eta_\mathrm{cal}(x_k, t_j)|^2
\end{equation}\end{small}additionally prevents large-amplitude samples from disproportionately influencing the loss and biasing the training process. 
Moreover, although advanced adaptive loss weighing schemes exist \cite[cf.][]{Chen2018, Heydari2019, Wang2020, Bischof2021}, we found that empirical manual tuning of the loss weights is sufficient for demonstrating the usefulness of our PINO.  Finally, the total loss (Eq. \ref{eq:L_total}) is minimized using the AdamW optimizer \cite{Kingma2017, Loshchilov2019} with an initial learning rate of $lr=0.001$, which is halved every 25 epochs. A batch size of 8 is used during training performed on a NVIDIA GeForce RTX 3090 GPU, with early stopping implemented after 50 epochs of no further improvement on the validation set.

\subsubsection{Error metric}
To evaluate the PINO's accuracy in reconstructing wave elevations or surface potentials after finishing the training, we employ the Surface Similarity Parameter (SSP) \cite{Perlin2014} as error metric. For two  spatio-temporal surfaces, $\mathbf{y}_\mathrm{HOSM}$ ($\eta_\mathrm{HOSM}$ or $\Phi^\mathrm{s}_\mathrm{HOSM}$) and $\tilde{\mathbf{y}}$ ($\tilde{\eta}$ or $\tilde{\Phi}^\mathrm{s}$) the $\mathrm{SSP}\in [0, 1]$ integrates deviations in phase, frequency and amplitude into a scalar measure by comparing their discrete Fourier transforms $F_{\mathbf{y}_\mathrm{HOSM}}$ and $F_{\tilde{\mathbf{y}}}$ as
\begin{small}\begin{equation}
    \mathrm{SSP} = \frac{\sqrt{\int | F_{\mathbf{y}_\mathrm{HOSM}}(\omega, k)-F_{\tilde{\mathbf{y}}}(\omega, k)|^2 d \omega d k}}{\sqrt{\int | F_{\mathbf{y}_\mathrm{HOSM}}(\omega, k)|^2 d \omega d k}+\sqrt{\int |F_{\tilde{\mathbf{y}}}(\omega, k)|^2 d \omega d k}}.
    \label{eq:SSP}
\end{equation}\end{small}Here, $\omega$ denotes the wave frequency and $k$ the wavenumber vector. The SSP recently has been applied in research related to ocean wave dynamics\cite{Wedler2023, Desmars2023, Kim2023, Ehlers2023, Ehlers2025, Dermatis2025, vanEssen2021} due to the straightforward error assessment, with $\mathrm{SSP}=0$ indicating perfect alignment and $\mathrm{SSP}=1$ indicating phase-inverted signals or a comparison against zero.

\section{Results}

The following sections present the results of training the PINO to reconstruct fully resolved wave surfaces $\tilde{\eta} \in \mathbb{R}^{n_x \times n_t}$ from sparse measurements. Sec. \ref{ch:results_buoy} examines the reconstruction from buoy measurements $\eta_\mathrm{m} \in \mathbb{R}^{n_\mathrm{buoy} \times n_t }$, denoted as mapping $\mathcal{M}_A:\eta_\mathrm{m}\rightarrow\tilde{\eta}$, while Sec. \ref{ch:results_radar} addresses reconstruction from radar snapshots $\xi_\mathrm{m} \in \mathbb{R}^{n_x \times n_\mathrm{snap}}$ as $\mathcal{M}_B:\xi_\mathrm{m}\rightarrow\tilde{\eta}$. Both models are trained using the sparse input data only, without access to a ground truth. The relatively small training set contains 634 sparse buoy $\eta_\mathrm{m}$ (case A) or radar $\xi_\mathrm{m}$ (case B) samples generated according to Section \ref{sec:data generation}. Performance is monitored on a validation set of 211 samples during training, and final evaluation is conducted in inference mode on an independent test set of another 211 samples.

\subsection{Reconstruction from buoy measurements}
\label{ch:results_buoy}
Case A in Fig. \ref{fig:PINO} investigates the mapping $\mathcal{M}_A: \eta_\mathrm{m} \rightarrow \tilde{\eta}$, where fully resolved wave surfaces are reconstructed from sparse buoy measurements. We assume, five buoys are positioned equidistantly within the region unaffected by the Tukey window tapering (see Sec. \ref{sec:loss and training}), resulting in positions $x_\mathrm{b}=\{488.76, \, 731.18 ,  \, 973.61 , \, 1216.03 , \, 1462.37 \}\, \mathrm{m}$. The PINO configuration is derived from a hyperparameter study (appendix Tab. \ref{tab:hyperparameters_caseA}) and employs $n_\mathrm{f} = 3$ Fourier layers, $n_\mathrm{m} = 128$ modes, and a latent width of $n_\mathrm{w} = 32$. The model $\mathcal{M}_A$ achieved minimum validation loss after 52 epochs, and the corresponding model weights are reloaded and used for evaluation on the test set.

\begin{figure*}
\centering
\includegraphics[scale=0.95]{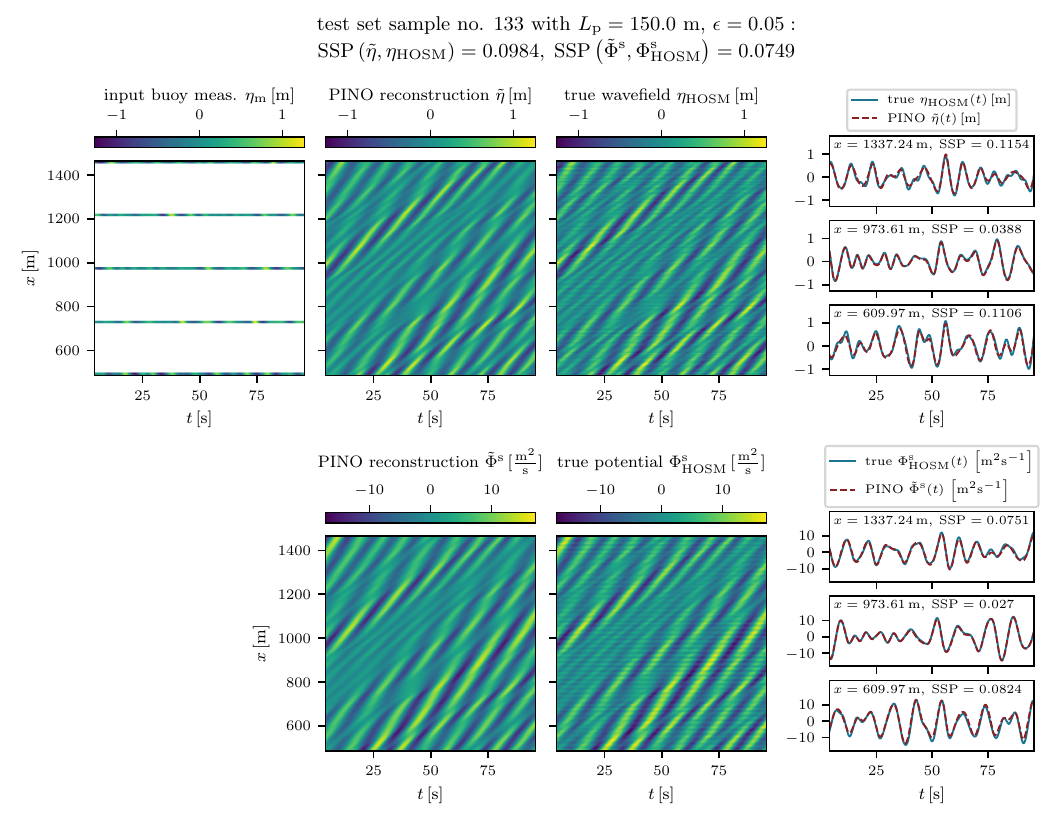}
\vspace{-0.3cm}
\caption{Reconstruction of the fully resolved wave surface $\tilde{\eta}$ and surface velocity potential $\tilde{\Phi}^\mathrm{s}$ from five sparse buoy measurements $\eta_\mathrm{m}$ using the PINO model ($\mathcal{M}_A$) for one test set sample. In the upper panels, $\tilde{\eta}$ and the reference surface $\eta_\mathrm{HOSM}$ show a good agreement. The same applies for $\tilde{\Phi}^\mathrm{s}$ and $\Phi^\mathrm{s}_\mathrm{HOSM}$ in the lower panels. Notably, reconstructions were achieved without access to fully resolved data during training of $\mathcal{M}_A$ - the model relies solely on the physics-based loss for reconstruction between buoys.}
\label{fig:buoy_samp_caseA}
\end{figure*}

Figure \ref{fig:buoy_samp_caseA} illustrates the performance of the trained PINO $\mathcal{M}_A$ for one representative test set sample (no. 133, $L_\mathrm{p}=150 \, \mathrm{m}$ and $\epsilon=0.05$). The upper left panel shows the five sparse input buoy measurements $\eta_\mathrm{m}$ from a top-down view, restricted to the region unaffected by Tukey window tapering. 
The subsequent panel displays the wave surface $\tilde{\eta}$ reconstructed by $\mathcal{M}_A$ in only $0.014\, \mathrm{s}$. It is followed by the fully resolved reference $\eta_\mathrm{HOSM}$, which is available only due to the synthetic nature of our data. Although never used during training, these references enable quantitative post-training validation of our developed method. From this, we observe that the reconstruction closely matches the reference, with a low error of $\mathrm{SSP} = 0.0984$. In the right panel, cross-sections at three spatial locations further confirm the accuracy: the central slice, which is aligned with a measurement location, shows near-perfect agreement between $\tilde{\eta}$ and $\eta_\mathrm{HOSM}$. Also the off-center slices, which are in between two measurement points, reveal only minor discrepancies in small-scale features, while the more relevant wave crests and troughs are largely well reconstructed. 
The bottom row of Fig. \ref{fig:buoy_samp_caseA} presents the simultaneously reconstructed surface velocity potential $\tilde{\Phi}^\mathrm{s}$ which also shows high agreement to its reference $\Phi^\mathrm{s}_\mathrm{HOSM}$, as reflected by a $\mathrm{SSP} = 0.0749$.

\begin{figure}[h]
\centering
\includegraphics[scale=0.95]{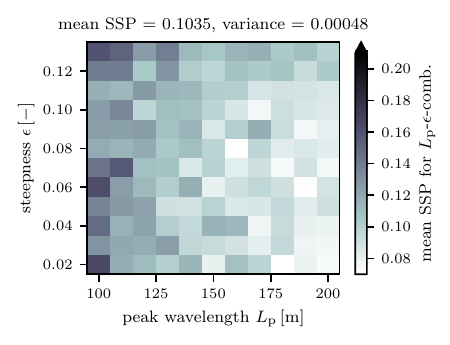}
\vspace{-0.4cm}
\caption{Mean wave surface reconstruction error $\mathrm{SSP}\left(\tilde{\eta}, \eta_\mathrm{HOSM}\right)$ across the test set using PINO $\mathcal{M}_A$ with buoy input data. Each cell shows the average SSP error for a specific combination of peak wavelength ($L_\mathrm{p}$) and wave steepness ($\epsilon$).  Comparatively low errors are observed for longer $L_\mathrm{p}$ and lower $\epsilon$, while accuracy remains satisfactory even for challenging parameter combinations.}
\label{fig:eroor_surface_caseA}
\end{figure}

After evaluating this individual sample, the performance of PINO $\mathcal{M}_A$ on the entire test set is assessed, which contains 210 additional buoy measurements with different $L_\mathrm{p}$- and $\epsilon$-values, and random phases. This analysis is summarized in Fig. \ref{fig:eroor_surface_caseA}, where each cell represents the mean SSP for a specific $L_\mathrm{p}$-$\epsilon$-combination, while the mean SSP across all test samples is $0.1035$. The figure illustrates that reconstruction accuracy slightly improves for wavefields with lower steepness ($\epsilon$) or longer peak wavelengths ($L_\mathrm{p}$). This observation may be attributed to multiple factors: (1)  Steeper, higher waves inherently exhibit stronger nonlinearities, complicating the reconstruction. (2) Samples for higher wave heights include smaller-scale wave components, as we are considering highly realistic sea states. This results in an underrepresentation of higher wave components in the training data, reducing model performance for high-$\epsilon$ samples. (3) Waves with shorter $L_\mathrm{p}$ may be more difficult to reconstruct due to the fixed buoy spacing of $\Delta x_\mathrm{b} \approx 246 \, \mathrm{m}$. While the longest waves in the dataset travel just slightly over one peak wavelength between measurement points, shorter waves undergo more rapid spatial variation between buoys. As a result, some features of short waves may be lost, leading to reduced reconstruction accuracy. However, since we aim for a model that is transferable to real ocean conditions, we must assume a uniform buoy spacing for all wavelengths. Nevertheless, even for short $L_\mathrm{p}$ and high $\epsilon$, the reconstruction errors remain within a satisfactory range, so that it can still be concluded that model $\mathcal{M}_A$ is capable of successfully generalizing its reconstruction capability across a range of wave conditions.

Overall, the results presented in Fig. \ref{fig:buoy_samp_caseA} provide strong evidence in support of hypothesis (\ref{hyp:1}a), as the PINO $\mathcal{M}_A$ successfully reconstructs spatio-temporal wave surfaces from sparse buoy measurements. Notably, ground truth data ($\eta_\mathrm{HOSM}$ and $\Phi^\mathrm{s}_\mathrm{HOSM}$) is used exclusively for post-training evaluation in this academic setting, and is not required during training. This directly supports hypothesis (\ref{hyp:2}), as the physics-informed training approach would enable learning from sparse, real-world buoy measurements, where high-resolution reference data is typically unavailable. This presents a situation where supervised learning is not applicable.
Furthermore, the PINO $\mathcal{M}_A$ achieves a reconstruction time of approximately $0.014\, \mathrm{s}$ per sample on the utilized standard workstation, which is comparable to the method of Köllisch et al. (2018) \cite{Koellisch2018}.  This rapid inference highlights the suitability of $\mathcal{M}_A$ for real-time wavefield prediction after the reconstruction, thereby supporting hypothesis (\ref{hyp:3}). Finally, as shown in Fig. \ref{fig:eroor_surface_caseA}, the PINO $\mathcal{M}_A$ demonstrates generalization across a broad range of wave conditions - a capability that standard PINN approaches do not achieve. Thus, the findings in this section also support hypothesis (\ref{hyp:4}).

\subsection{Reconstruction from radar measurements}
\label{ch:results_radar}

Building upon the successful reconstruction of wavefields from sparse buoy measurements, we now address case B in Fig. \ref{fig:PINO}, which focuses on reconstructing phase-resolved wave surfaces from radar snapshot data, formally denoted as $\mathcal{M}_B: \xi_\mathrm{m}\rightarrow \tilde{\eta}$. Unlike buoy measurements, radar intensities do not exhibit a direct proportionality to wave elevations, thereby introducing additional complexity to the reconstruction task. To address this, we incorporate a single calibration buoy measurement $\eta_\mathrm{cal}$ in the sensor loss term $\mathcal{L}_\mathrm{sensor}$, (but not as a direct input) to enable the model to learn the mapping between radar intensity values and wave elevation values.
For this task, a slightly modified PINO architecture yielded the best performance in a hyperparameter study (appendix Tab. \ref{tab:hyperparameters_caseB}). This configuration employs $n_\mathrm{f}=3$ Fourier layers and $n_\mathrm{m}=128$ modes, but a larger latent width of $n_\mathrm{w} = 64$, reflecting the increased complexity of the problem arising from the indirect nature of the measurements. The model $\mathcal{M}_B$ achieved its lowest validation loss after 243 epochs, at which point the corresponding weights are reloaded.

\begin{figure*}
\centering
\includegraphics[scale=0.95]{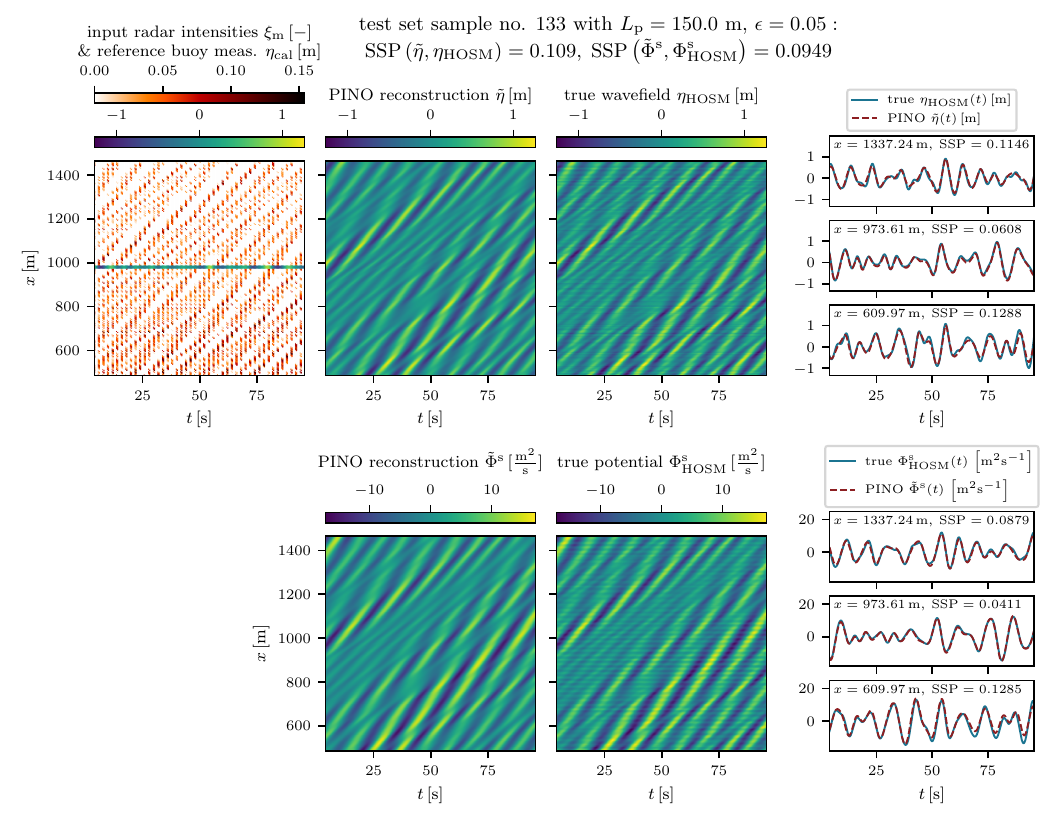}
\vspace{-0.3cm}
\caption{Reconstruction of a fully resolved wave surface $\tilde{\eta}$ and surface velocity potential $\tilde{\Phi}^\mathrm{s}$ from radar snapshot intensities $\xi_\mathrm{m}$ using the trained PINO model ($\mathcal{M}_B$) for the same test sample as in Fig. \ref{fig:buoy_samp_caseA}. The radar input snapshots in the left panel (with $\Delta t_\mathrm{r} = 2 \, \mathrm{s}$) exhibit regions of zero intensity due to shadowing modulation. The location of the calibration buoy measurement $\eta_\mathrm{cal}$ is also indicated. Despite the increased complexity of radar-based reconstruction, that cause a small decrease in accuracy compared to the previous case, $\tilde{\eta}$ and $\tilde{\Phi}^\mathrm{s}$ show a satisfying agreement with the references $\eta_\mathrm{HOSM}$ and $\Phi^\mathrm{s}_\mathrm{HOSM}$.  }
\label{fig:radar_samp_caseB}
\end{figure*}

For the evaluation of PINO $\mathcal{M}_B$, we employ the same test set sample (no. 133) as in the previous section. The input sequence of radar intensity snapshots $\xi_\mathrm{m}$ is displayed in the left panel of Fig. \ref{fig:radar_samp_caseB}. These snapshots, recorded with $\Delta t_\mathrm{r}=2\, \mathrm{s}$, reveal certain spatial regions impacted by shadowing modulation, evident as zero-intensities.
The same panel also indicates the location of the calibration buoy measurement $\eta_\mathrm{cal}$ at $x_\mathrm{b}=973.61\, \mathrm{m}$, which only serves for calibration purposes of $\mathcal{L}_\mathrm{sensor}$ during training but is not required during inference.
The adjacent panels present the reconstructed wave surface $\tilde{\eta}$ and the corresponding reference field $\eta_\mathrm{HOSM}$. As in the previous case, reference fields in this form were not available during training. Although the reconstruction error ($\mathrm{SSP}=0.1090$) is marginally higher than in case A, the reconstruction $\tilde{\eta}$ demonstrates strong agreement with the reference $\eta_\mathrm{HOSM}$, as further confirmed by the cross-sections in the right panel. Similar accuracy is observed for the surface velocity potential in the lower panels, where the PINO reconstruction $\tilde{\Phi}^\mathrm{s}$ achieves a $\mathrm{SSP}=0.0949$ relative to the reference $\Phi^\mathrm{s}_\mathrm{HOSM}$.

\begin{figure}
\centering
\includegraphics[scale=0.95]{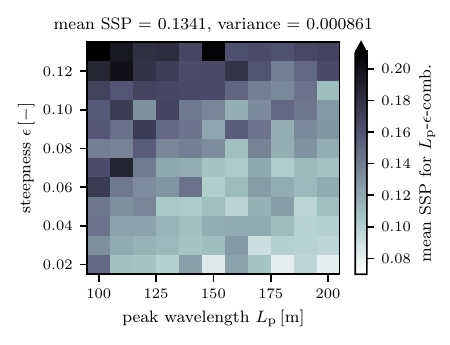}
\vspace{-0.4cm}
\caption{Mean wave surface reconstruction error $\mathrm{SSP}\left(\tilde{\eta}, \eta_\mathrm{HOSM}\right)$ across the test set using PINO $\mathcal{M}_B$ with radar snapshot inputs $\xi_\mathrm{m}$. Each cell shows the average SSP error for a specific $L_\mathrm{p}$-$\epsilon$-combination. Compared to buoy-based reconstructions in Fig. \ref{fig:eroor_surface_caseA}, errors are generally higher, particularly for samples with high $\epsilon$, due to increased shadowing effects in $\xi_\mathrm{m}$.}
\vspace{-0.3cm}
\label{fig:eroor_surface_caseB}
\end{figure}

As shown in Fig. \ref{fig:eroor_surface_caseB}, the average reconstruction error for wave surfaces from radar measurements across the entire test dataset is $\mathrm{SSP} = 0.1341$, which is higher than the error observed for buoy-based reconstructions ($\mathrm{SSP}=0.1035$; see Fig. \ref{fig:eroor_surface_caseA}), but remains within an satisfying range. As before, each cell of the figure represents the mean SSP for all test set samples corresponding to a specific $L_\mathrm{p}$-$\epsilon$ parameter combination. Notably, in contrast to the previous section, samples with higher steepness now exhibit a more pronounced increase in reconstruction error. This increase cannot be attributed solely to wave nonlinearity or the underrepresentation of higher wave components in the training data, as was previously discussed. Instead, it is primarily a consequence of geometric shadowing modulation effects: samples with higher steepness are more severely affected by shadowing modulation, resulting in radar measurements that inherently contain less information, making the reconstruction task more challenging for these cases.

\begin{figure*}
\centering
\includegraphics[scale=0.95]{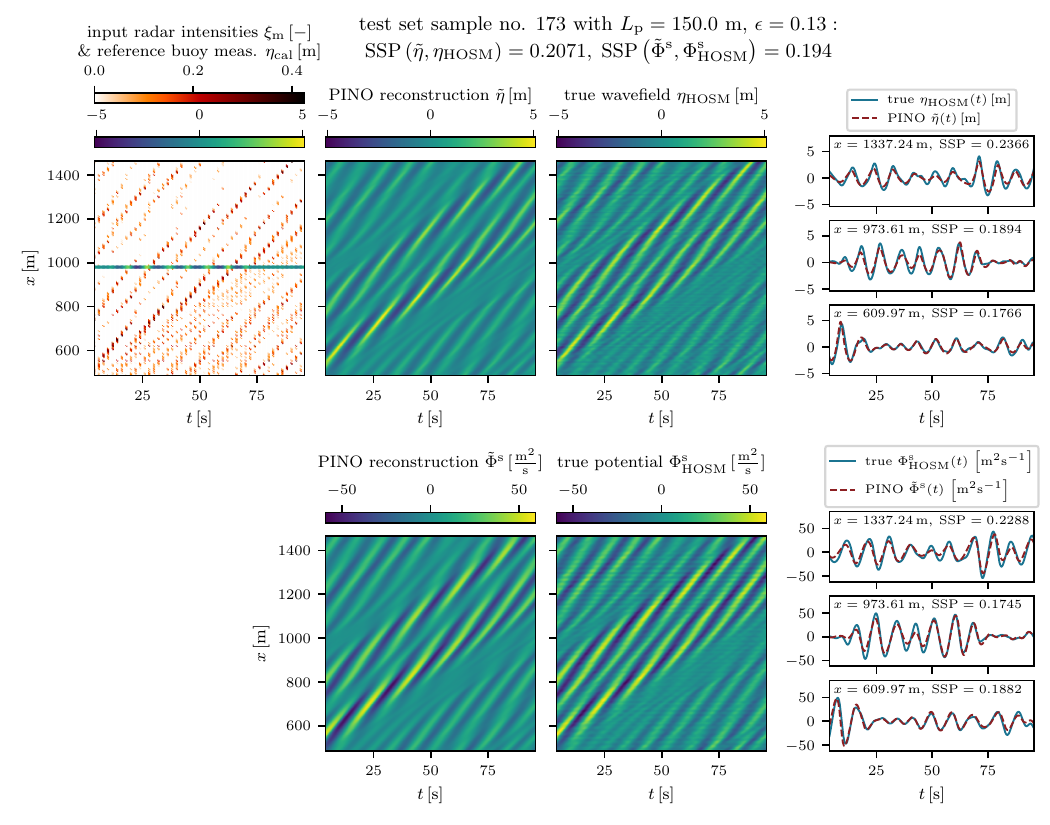}
\vspace{-0.3cm}
\caption{Reconstruction of a fully resolved wave surface $\tilde{\eta}$ and surface velocity potential $\tilde{\Phi}^\mathrm{s}$ from radar snapshot intensities $\xi_\mathrm{m}$ using the trained PINO model ($\mathcal{M}_B$) for another same test sample with higher steepness $\epsilon=0.13$ (previously $\epsilon=0.05$ in Fig. \ref{fig:radar_samp_caseB}). Radar snapshots show extensive regions of zero intensity due to pronounced shadowing effects, resulting in reduced information density compared to lower-$\epsilon$ cases. Despite increased reconstruction errors, especially at greater distances from the radar (at $x=0\, \mathrm{m}$), the predicted fields $\tilde{\eta}$ and $\tilde{\Phi}^\mathrm{s}$ remain in good qualitative agreement with the references $\eta_\mathrm{HOSM}$ and $\Phi^\mathrm{s}_\mathrm{HOSM}$. }
\label{fig:radar_samp_caseB-2}
\end{figure*}

This relationship between reduced information density in the radar input $\xi_\mathrm{m}$ and increased reconstruction errors is further illustrated in Fig. \ref{fig:radar_samp_caseB-2}, which presents results for another test set sample (no. 173) characterized by $L_\mathrm{p}=150, \mathrm{m}$ and the maximum steepness $\epsilon=0.13$. The left panel shows that a significantly larger portion of the input exhibits zero intensity values (in contrast to sample no. 133 with $\epsilon=0.05$ in Fig.~\ref{fig:radar_samp_caseB}).
This pronounced shadowing effect directly impacts the reconstruction quality of $\tilde{\eta}$, particularly at greater distances from the radar located at $x=0\,\mathrm{m}$.  This trend is evident in the cross-section at $x=1337.24\,\mathrm{m}$, where the SSP is higher compared to the closer cross-section locations.
Nevertheless, considering that sample no. 173 exhibits one of the highest reconstruction errors in the entire test dataset, the accuracy of $\tilde{\eta}$ and $\tilde{\Phi}^\mathrm{s}$ remains satisfactory when compared to their references. While some wave crests and troughs are not perfectly captured, the phase alignment is highly accurate, and the qualitative agreement in the top-view plots is still acceptable. Moreover, the observation of increased reconstruction errors from radar data for higher $\epsilon$-values is consistent with findings previously reported for the purely supervised approach in Ehlers et al. (2023) \cite{Ehlers2023} and can therefore not be attributed to the underlying principle of the PINO approach but is rather a physical issue. \\

In summary, the results of this section demonstrate that the PINO model $\mathcal{M}_B$ can successfully reconstruct phase-resolved nonlinear wave surfaces from radar snapshot measurements, supporting hypothesis (\ref{hyp:1}b). While reconstruction errors are higher than for buoy-based inputs in the last section, primarily due to information loss from shadowing modulation for high steepness samples, the accuracy remains robust across a broad range of conditions. As with the buoy case, no ground truth wave fields are needed during training, supporting hypothesis (\ref{hyp:2}). Moreover, inference times remain suitable for real-time applications (\ref{hyp:3}). Although the radar-based PINO model ($\mathcal{M}_\mathrm{B}$) does not generalize to a wide range of wave conditions as effectively as the buoy-based model ($\mathcal{M}_A$), its limitations are primarily caused by the lower information content and measurement artefacts present in radar data, rather than by the model architecture itself.  Notably, the generalisation performance of $\mathcal{M}_B$ still remains substantially better than that of conventional PINN methods, which show no generalization capability at all. Therefore, despite the challenges posed by radar measurements, hypothesis (\ref{hyp:4}) is still supported, confirming the potential of PINOs for practical wavefield monitoring using diverse sensor types.


\section{Conclusion and Outlook}
\label{ch:conclusion}

In this study, we presented a Physics-Informed Neural Operator (PINO) tailored for wave data assimilation. Our PINO reconstructs fully-resolved free-surface wave elevation fields and corresponding velocity potential fields from sparse measurements, such as buoy data (case A) or radar snapshots combined with data from a single calibration buoy (case B). A key advantage of our approach is that it eliminates the need for fully-resolved ground-truth data during training, which is typically impractical to obtain in real-ocean applications. This implies, instead of directly minimizing the reconstruction error relative to a ground truth, as in conventional supervised learning, the PINO learns to generate physically consistent wave fields by minimizing residuals of the governing equations. Specifically, we incorporate the residuals of the free-surface boundary conditions, derived from the potential flow theory of ocean gravity waves, directly into the loss function. These physics-based constraints guide the model to infer consistent wave fields between sparsely distributed measurement locations.

Compared to our previous methods for wave surface reconstruction, the proposed PINO combines the strengths of our FNO-based approach \cite{Ehlers2023} and PINN-based approach\cite{Ehlers2025}, while addressing their limitations.
First, the PINO is highly data-efficient as a training set of only 634 sparse input samples is enough to achieve a satisfying performance on the test set, quantified by mean SSP values of $\mathrm{SSP}=0.1035$ (case A) and $\mathrm{SSP}=0.1341$ (case B). This demonstrates that embedding physical knowledge significantly reduces the amount of data required, compared to conventional machine learning models such as our FNO-approach\cite{Ehlers2023}, which needed approximately three times more training data to reach similar accuracy. Moreover, unlike the FNO \cite{Ehlers2023}, the PINO does not rely on labelled datasets of ground-truth wave elevations, making it directly applicable in real-world scenarios with only sparse observational data.  In contrast to our PINN-approach \cite{Ehlers2025}, the PINO can assimilate indirect measurements such as radar backscatter and does not require a time-consuming solution of a separate inverse problem for each wave instance. Instead, once trained, the PINO generalizes across wave conditions and enables rapid, instance-independent inference, taking by far less than one second to reconstruct a previously unseen wave instance.
Overall, these properties make the proposed PINO approach a promising candidate for real-time, physics-constrained wave field reconstruction from sparse and indirect measurements, with high potential for operational application in engineering and oceanography.

Currently, the proposed PINO approach has been demonstrated on long-crested, unidirectional ($\mathrm{1D}+t$) wave surfaces. Extending this framework to realistic two-dimensional ($\mathrm{2D}+t$) wave fields, where waves arrive from multiple directions and interact through complex superposition, is the future perspective for this work. Since the core of the PINO relies on Fourier layers that can be extended to higher dimensions, this extension does not present fundamental methodological challenges. However, the increased dimensionality will demand substantially more computational resources or efficient parallelization strategies. Although the more complex wave fields may slightly reduce reconstruction accuracy, we expect that the embedded physics constraints will sustain robust performance. Importantly, once trained, the PINO should still enable extremely fast inference, making real-time physics-informed wave field reconstruction in $\mathrm{2D}+t$ scenarios feasible for practical applications.

\vfill

\section*{Funding}
This work was supported by the Deutsche Forschungsgemeinschaft (DFG - German Research Foundation) [project number 277972093: Excitability of Ocean Rogue Waves]. 

\section*{Declaration of interests}
The authors declare that they have no known competing financial interests or personal relationships that could have appeared to influence the work reported in this paper.

\section*{Declaration of generative AI and AI-
assisted technologies in the writing process}

The investigations were conducted and the manuscript was written completely by the authors. Once the manuscript was completed, the authors used ChatGPT 3.5 (\url{https://chat.openai.com/}) in order to improve its grammar and readability. After using this tool, the authors reviewed and edited the content as needed and take full responsibility for the content of the publication.

\section*{Data Availability Statement}
The data that supports the findings of this study are available from the corresponding author upon reasonable request.

\appendix
\section{Hyperparameter study}

To identify the optimal configuration for the PINO architecture shown in Fig. \ref{fig:PINO}, a hyperparameter study was conducted for both, case A (reconstruction of wave elevations $\tilde{\eta}$ from buoy measurements $\eta_\mathrm{m}$) and case B (reconstruction of wave fields $\tilde{\eta}$  from radar snapshots $\xi_\mathrm{m}$). In this hyperparameter study, the number of FNO layers $n_\mathrm{F}$, the number of Fourier modes retained without truncation $n_\mathrm{m}$, and the width of the latent representation $n_\mathrm{w}$, were systematically varied. Additionally, the total number of resulting weights from all layers ($I$, $P$, $R_{\mathrm{L}_i}$, $W_{\mathrm{L}_i}$, $Q$) are counted.

Table \ref{tab:hyperparameters_caseA} clearly shows that for case A, a configuration with $n_\mathrm{F}=3$, $n_\mathrm{m}=128$ and $n_\mathrm{w}=32$ performed best, achieving the lowest loss $\mathcal{L}$ and SSP values on the validation set. Additionally, the variance of the SSP values for this configuration was among the lowest.

In contrast, for case B in Table \ref{tab:hyperparameters_caseB}, due to the increased complexity of radar-based reconstructions, the best results were achieved with a larger latent dimension $n_\mathrm{w}=64$, while maintaining $n_\mathrm{F}=3$ and $\mathrm{n}_\mathrm{m}=128$.

For the rows in both tables marked with an underscore (-), training with a batch size of 8 was not possible on the available hardware (NVIDIA GeForce RTX 3090 GPU). However, it can be expected that even better reconstruction results could be achieved with more powerful hardware.

\begin{table*}[!htp]
\footnotesize
\centering
\caption{Results of the hyperparameter study for the PINO architecture for the reconstruction of wave fields from buoy measurements ($\mathcal{M}_A: \eta_\mathrm{m}\rightarrow\tilde{\eta}$) in Sec. \ref{ch:results_buoy}. Each investigated architecture is characterized by several Fourier layers $n_\mathrm{F}$, a width of the latent representation $n_\mathrm{w}$ and the modes $n_\mathrm{m}$ for truncation of the layers' Fourier series.}
\begin{tabular}{cccrcccccccc}
\toprule
\toprule
\multicolumn{3}{c}{PINO hyperparameters} & \multicolumn{2}{c}{} & \multicolumn{2}{c}{loss $\mathcal{L}$} & & \multicolumn{3}{c}{SSP} \\
\cmidrule(rl){1-3} \cmidrule(rl){6-7} \cmidrule(rl){9-11} 
layers $n_\mathrm{F}$ & modes $n_\mathrm{m}$ & width $n_\mathrm{w}$ & \#weights & \hspace{0.2cm}epochs \hspace{0.2cm} & train & val && train & val & val variance  \\ 
\midrule
\midrule

\multirow{4}{*}{2} & \multirow{4}{*}{64} 
      & 16 & 8,398,521   & 249 & 0.04145 & 0.04466 && 0.1287 & 0.1283 & 0.001061\\
    & & 32 & 33,567,993  & 235 & 0.03328 & 0.03885 && 0.1163 & 0.1161 & 0.000824\\
    & & 48 & 75,515,705  & 223 & 0.03115 & 0.03724 && 0.1127 & 0.1126 & 0.000746\\
    & & 64 & 134,241,657 & 195 & 0.02896 & 0.03646 && 0.1102 & 0.1105 & 0.000703\vspace{0.12cm}\\

\multirow{4}{*}{2} & \multirow{4}{*}{128} 
      & 16 & 33,564,345  & 243 & 0.03551 & 0.03956 && 0.1169 & 0.1164 & 0.000718\\
    & & 32 & 134,231,289 & 153 & 0.03146 & 0.03850 && 0.1101 & 0.1099 & 0.000586\\
    & & 48 & 302,008,121 & 203 & 0.02844 & 0.03635 && 0.1053 & 0.1056 & 0.000503\\
    & & 64 & 536,894,841 & 180 & 0.02862 & 0.03567 && 0.1058 & 0.1064 & 0.000511\vspace{0.12cm}\\

\multirow{4}{*}{2} & \multirow{4}{*}{256} 
      & 16 &  134,227,641 & 218 & 0.03542 & 0.03984 && 0.1181 & 0.1179 & 0.000743\\
    & & 32 &  536,884,473 & 193 & 0.02948 & 0.03725 && 0.1073 & 0.1073 & 0.000562\\
    & & 48 &  1,207,977,785 & - & - & - &  & - & - & -\\
    & & 64 &  2,147,507,577 & - & - & - &  & - & - & -\vspace{0.12cm}\\

\midrule
 
\multirow{4}{*}{3} & \multirow{4}{*}{64} 
      & 16 &  12,593,097  & 208 & 0.03194 & 0.03898 && 0.1139 & 0.1141 & 0.000810\\
    & & 32 &  50,346,265  &  84 & 0.02856 & 0.03841 && 0.1089 & 0.1093 & 0.000706\\
    & & 48 &  113,266,793 &  50 & 0.03082 & 0.03730 && 0.1111 & 0.1112 & 0.000729\\
    & & 64 &  201,354,681 &  77 & 0.02854 & 0.03710 && 0.1086 & 0.1089 & 0.000680\vspace{0.12cm}\\

\multirow{4}{*}{3} & \multirow{4}{*}{128} 
      & 16 &  50,341,833  & 84 & 0.03151 & 0.03911 && 0.1087 & 0.1084 & 0.000593\\
    & & 32 & \cellcolor{DynCyan!40}201,341,209 & \cellcolor{DynCyan!40}52 & \cellcolor{DynCyan!40}0.02954 & \cellcolor{DynCyan!40}0.03803 && \cellcolor{DynCyan!40}0.1048 & \cellcolor{DynCyan!40}0.1049 & \cellcolor{DynCyan!40}0.000501\\
    & & 48 &  453,005,417 & 52 & 0.03038 & 0.03731 && 0.1063 & 0.1059 & 0.000501\\
    & & 64 &  805,334,457 & 50 & 0.03129 & 0.03675 && 0.1065 & 0.1068 & 0.000520\vspace{0.12cm}\\

\multirow{4}{*}{3} & \multirow{4}{*}{256} 
      & 16 &  201,336,777   & 76 & 0.03006 & 0.03844 && 0.1065 & 0.1073 & 0.000544\\
    & & 32 &  805,320,985   & 50 & 0.02993 & 0.03715 && 0.1060 & 0.1058 & 0.000534\\
    & & 48 &  1,811,959,913 & - & - & - &  & - & - & -\\
    & & 64 &  3,221,253,561 & - & - & - &  & - & - & -\vspace{0.12cm}\\

\midrule
   
\multirow{4}{*}{4} & \multirow{4}{*}{64} 
      & 16 &  16,787,673  & 76 & 0.03033 & 0.04025 && 0.1121 & 0.1135 & 0.000801\\
    & & 32 &  67,124,537  & 54 & 0.02636 & 0.03933 && 0.1069 & 0.1079 & 0.000689\\
    & & 48 &  151,017,881 & 50 & 0.02859 & 0.03688 && 0.1081 & 0.1085 & 0.000677\\
    & & 64 &  268,467,705 & 51 & 0.02980 & 0.03762 && 0.1087 & 0.1088 & 0.000698\vspace{0.12cm}\\

\multirow{4}{*}{4} & \multirow{4}{*}{128} 
      & 16 &   67,119,321 & 50 & 0.02982 & 0.03883 && 0.1054 & 0.1055 & 0.000517\\
    & & 32 &  268,451,129 & 26 & 0.03101 & 0.03773 && 0.1053 & 0.1054 & 0.000531\\
    & & 48 &  604,002,713 & 30 & 0.02935 & 0.03721 && 0.1045 & 0.1053 & 0.000576\\
    & & 64 &  1,073,774,073 & - & - & - &  & - & - & - \vspace{0.12cm}\\

\multirow{4}{*}{4} & \multirow{4}{*}{256} 
      & 16 &  268,445,913 & 51 & 0.02912 & 0.0384 && 0.1053 & 0.1057 & 0.000498\\
    & & 32 &  1,073,757,497 & - & - & - &  & - & - & -\\
    & & 48 &  2,415,942,041 & - & - & - &  & - & - & -\\
    & & 64 & 4,294,999,545 & - & - & - &  & - & - & -\vspace{0.12cm}\\
\bottomrule
\bottomrule
\end{tabular}
\label{tab:hyperparameters_caseA}
\vspace{1.2cm}
\end{table*}

\begin{table*}[!htp]
\centering
\footnotesize
\caption{Results of the hyperparameter study for the PINO architecture for the reconstruction of wave fields from radar snapshot measurements ($\mathcal{M}_B: \xi_\mathrm{m}\rightarrow\tilde{\eta}$) in Sec. \ref{ch:results_radar}. Each investigated architecture is characterized by several Fourier layers $n_\mathrm{F}$, a width of the latent representation $n_\mathrm{w}$ and the modes $n_\mathrm{m}$ for truncation of the layers' Fourier series.}
\begin{tabular}{cccrcccccccc}
\toprule
\toprule
\multicolumn{3}{c}{PINO hyperparameters} & \multicolumn{2}{c}{} & \multicolumn{2}{c}{loss $\mathcal{L}$} & & \multicolumn{3}{c}{SSP} \\
\cmidrule(rl){1-3} \cmidrule(rl){6-7} \cmidrule(rl){9-11} 
layers $n_\mathrm{F}$ & modes $n_\mathrm{m}$ & width $n_\mathrm{w}$ & \#weights & \hspace{0.2cm}epochs \hspace{0.2cm} & train & val && train & val & val variance  \\ 
\midrule
\midrule

\multirow{4}{*}{2} & \multirow{4}{*}{64} 
      & 16 &  8,421,021 & 248 & 0.05935 & 0.08450 && 0.1779 & 0.1803 & 0.001837 \\
    & & 32 & 33,590,493 & 244 & 0.03649 & 0.07329 && 0.1690 & 0.1712 & 0.001609 \\
    & & 48 & 75,538,205 & 201 & 0.02511 & 0.07818 && 0.1624 & 0.1652 & 0.001520 \\
    & & 64 &134,264,157 &  98 & 0.03036 & 0.08295 && 0.1607 & 0.1645 & 0.001457\vspace{0.12cm}\\

\multirow{4}{*}{2} & \multirow{4}{*}{128} 
      & 16 &  33,586,845 & 243 & 0.03419 & 0.06904 && 0.1543 & 0.1571 & 0.001569 \\
    & & 32 & 134,253,789 & 247 & 0.01717 & 0.05807 && 0.1541 & 0.1564 & 0.001376 \\
    & & 48 & 302,030,621 & 246 & 0.01412 & 0.05522 && 0.1497 & 0.1522 & 0.001173 \\
    & & 64 & 536,917,341 & 246 & 0.01373 & 0.05618 && 0.1454 & 0.1480 & 0.001126\vspace{0.12cm}\\

\multirow{4}{*}{2} & \multirow{4}{*}{256} 
      & 16 & 134,250,141 & 249 & 0.02234 & 0.07399 && 0.1522 & 0.1547 & 0.001295\\
    & & 32 & 536,906,973 & 245 & 0.01793 & 0.06739 && 0.1654 & 0.1678 & 0.001551\\
    & & 48 &  & - & - & - &  & - & - & -\\
    & & 64 &  & - & - & - &  & - & - & -\vspace{0.12cm}\\

\midrule
 
\multirow{4}{*}{3} & \multirow{4}{*}{64} 
      & 16 & 12,615,597  & 247 & 0.03124 & 0.06374 && 0.1707 & 0.1727 & 0.001360\\
    & & 32 & 50,368,765  & 197 & 0.01399 & 0.06040 && 0.1553 & 0.1579 & 0.001136\\
    & & 48 & 113,289,293 & 242 & 0.01360 & 0.07132 && 0.1507 & 0.1542 & 0.001276\\
    & & 64 & 201,377,181 & 241 & 0.01179 & 0.06978 && 0.1514 & 0.1548 & 0.001226\vspace{0.12cm}\\

\multirow{4}{*}{3} & \multirow{4}{*}{128} 
      & 16 &  50,364,333 & 249 & 0.01649 & 0.04941 && 0.1465 & 0.1496 & 0.001276\\
    & & 32 & 201,363,709 & 238 & 0.01096 & 0.04317 && 0.1400 & 0.1425 & 0.001052\\
    & & 48 & 453,027,917 & 246 & 0.01032 & 0.04543 && 0.1401 & 0.1428 & 0.001071\\
    & & 64 & \cellcolor{DynCyan!40}805,356,957 & \cellcolor{DynCyan!40}243 & \cellcolor{DynCyan!40}0.00950 & \cellcolor{DynCyan!40}0.04437 && \cellcolor{DynCyan!40}0.1343 & \cellcolor{DynCyan!40}0.1369 & \cellcolor{DynCyan!40}0.000920\vspace{0.12cm}\\

\multirow{4}{*}{3} & \multirow{4}{*}{256} 
      & 16 & 201,359,277 & 243 & 0.01431 & 0.05628 && 0.1508 & 0.1537 & 0.001287\\
    & & 32 & 805,343,485 & 248 & 0.00934 & 0.03932 && 0.1437 & 0.1456 & 0.000839\\
    & & 48 &  & - & - & - &  & - & - & -\\
    & & 64 &  & - & - & - &  & - & - & -\vspace{0.12cm}\\

\midrule
   
\multirow{4}{*}{4} & \multirow{4}{*}{64} 
      & 16 &  16,810,173 & 235 & 0.02077 & 0.05630 && 0.1806 & 0.1829 & 0.001177\\
    & & 32 &  67,147,037 &  97 & 0.01166 & 0.05507 && 0.1598 & 0.1628 & 0.001072\\
    & & 48 & 151,040,381 & 224 & 0.00846 & 0.05519 && 0.1638 & 0.1670 & 0.001214\\
    & & 64 & 268,490,205 & 121 & 0.00895 & 0.05676 && 0.1628 & 0.1671 & 0.00127\vspace{0.12cm}\\

\multirow{4}{*}{4} & \multirow{4}{*}{128} 
      & 16 &  67,141,821 & 248 & 0.01001 & 0.04011 && 0.1412 & 0.1446 & 0.001013\\
    & & 32 & 268,473,629 & 239 & 0.00804 & 0.03744 && 0.1504 & 0.1529 & 0.000922\\
    & & 48 &  & - & - & - && - & - & -\\
    & & 64 &  & - & - & - && - & - & -\vspace{0.12cm}\\

\multirow{4}{*}{4} & \multirow{4}{*}{256} 
      & 16 & 268,468,413 & 249 & 0.00915 & 0.03916 && 0.1478 & 0.1506 & 0.000837\\
    & & 32 &  & - & - & - && - & - & -\\
    & & 48 &  & - & - & - && - & - & -\\
    & & 64 &  & - & - & - && - & - & -\vspace{0.12cm}\\
\bottomrule
\bottomrule
\end{tabular}
\label{tab:hyperparameters_caseB}
\vspace{1.2cm}
\end{table*}

\vspace{2cm}
\bibliography{aipsamp}

\providecommand{\noopsort}[1]{}\providecommand{\singleletter}[1]{#1}
\begin{thebibliography}{60}%
\makeatletter
\providecommand \@ifxundefined [1]{%
 \@ifx{#1\undefined}
}%
\providecommand \@ifnum [1]{%
 \ifnum #1\expandafter \@firstoftwo
 \else \expandafter \@secondoftwo
 \fi
}%
\providecommand \@ifx [1]{%
 \ifx #1\expandafter \@firstoftwo
 \else \expandafter \@secondoftwo
 \fi
}%
\providecommand \natexlab [1]{#1}%
\providecommand \enquote  [1]{``#1''}%
\providecommand \bibnamefont  [1]{#1}%
\providecommand \bibfnamefont [1]{#1}%
\providecommand \citenamefont [1]{#1}%
\providecommand \href@noop [0]{\@secondoftwo}%
\providecommand \href [0]{\begingroup \@sanitize@url \@href}%
\providecommand \@href[1]{\@@startlink{#1}\@@href}%
\providecommand \@@href[1]{\endgroup#1\@@endlink}%
\providecommand \@sanitize@url [0]{\catcode `\\12\catcode `\$12\catcode `\&12\catcode `\#12\catcode `\^12\catcode `\_12\catcode `\%12\relax}%
\providecommand \@@startlink[1]{}%
\providecommand \@@endlink[0]{}%
\providecommand \url  [0]{\begingroup\@sanitize@url \@url }%
\providecommand \@url [1]{\endgroup\@href {#1}{\urlprefix }}%
\providecommand \urlprefix  [0]{URL }%
\providecommand \Eprint [0]{\href }%
\providecommand \doibase [0]{http://dx.doi.org/}%
\providecommand \selectlanguage [0]{\@gobble}%
\providecommand \bibinfo  [0]{\@secondoftwo}%
\providecommand \bibfield  [0]{\@secondoftwo}%
\providecommand \translation [1]{[#1]}%
\providecommand \BibitemOpen [0]{}%
\providecommand \bibitemStop [0]{}%
\providecommand \bibitemNoStop [0]{.\EOS\space}%
\providecommand \EOS [0]{\spacefactor3000\relax}%
\providecommand \BibitemShut  [1]{\csname bibitem#1\endcsname}%
\let\auto@bib@innerbib\@empty
\bibitem [{\citenamefont {Engsig-Karup}, \citenamefont {Bingham},\ and\ \citenamefont {Lindberg}(2009)}]{Engsig2009}%
  \BibitemOpen
  \bibfield  {author} {\bibinfo {author} {\bibfnamefont {A.~P.}\ \bibnamefont {Engsig-Karup}}, \bibinfo {author} {\bibfnamefont {H.~B.}\ \bibnamefont {Bingham}}, \ and\ \bibinfo {author} {\bibfnamefont {O.}~\bibnamefont {Lindberg}},\ }\bibfield  {title} {\enquote {\bibinfo {title} {An efficient flexible-order model for 3d nonlinear water waves},}\ }\href@noop {} {\bibfield  {journal} {\bibinfo  {journal} {Journal of computational physics}\ }\textbf {\bibinfo {volume} {228}},\ \bibinfo {pages} {2100--2118} (\bibinfo {year} {2009})}\BibitemShut {NoStop}%
\bibitem [{\citenamefont {Wu}\ and\ \citenamefont {Taylor}(1994)}]{Wu1994}%
  \BibitemOpen
  \bibfield  {author} {\bibinfo {author} {\bibfnamefont {G.}~\bibnamefont {Wu}}\ and\ \bibinfo {author} {\bibfnamefont {R.~E.}\ \bibnamefont {Taylor}},\ }\bibfield  {title} {\enquote {\bibinfo {title} {Finite element analysis of two-dimensional non-linear transient water waves},}\ }\href@noop {} {\bibfield  {journal} {\bibinfo  {journal} {Applied Ocean Research}\ }\textbf {\bibinfo {volume} {16}},\ \bibinfo {pages} {363--372} (\bibinfo {year} {1994})}\BibitemShut {NoStop}%
\bibitem [{\citenamefont {Ma}\ and\ \citenamefont {Yan}(2006)}]{Ma2006}%
  \BibitemOpen
  \bibfield  {author} {\bibinfo {author} {\bibfnamefont {Q.}~\bibnamefont {Ma}}\ and\ \bibinfo {author} {\bibfnamefont {S.}~\bibnamefont {Yan}},\ }\bibfield  {title} {\enquote {\bibinfo {title} {Quasi ale finite element method for nonlinear water waves},}\ }\href@noop {} {\bibfield  {journal} {\bibinfo  {journal} {Journal of computational physics}\ }\textbf {\bibinfo {volume} {212}},\ \bibinfo {pages} {52--72} (\bibinfo {year} {2006})}\BibitemShut {NoStop}%
\bibitem [{\citenamefont {Grilli}, \citenamefont {Guyenne},\ and\ \citenamefont {Dias}(2001)}]{Grilli2001}%
  \BibitemOpen
  \bibfield  {author} {\bibinfo {author} {\bibfnamefont {S.~T.}\ \bibnamefont {Grilli}}, \bibinfo {author} {\bibfnamefont {P.}~\bibnamefont {Guyenne}}, \ and\ \bibinfo {author} {\bibfnamefont {F.}~\bibnamefont {Dias}},\ }\bibfield  {title} {\enquote {\bibinfo {title} {A fully non-linear model for three-dimensional overturning waves over an arbitrary bottom},}\ }\href@noop {} {\bibfield  {journal} {\bibinfo  {journal} {International journal for numerical methods in fluids}\ }\textbf {\bibinfo {volume} {35}},\ \bibinfo {pages} {829--867} (\bibinfo {year} {2001})}\BibitemShut {NoStop}%
\bibitem [{\citenamefont {Dommermuth}\ and\ \citenamefont {Yue}(1987)}]{Dommermuth1987}%
  \BibitemOpen
  \bibfield  {author} {\bibinfo {author} {\bibfnamefont {D.~G.}\ \bibnamefont {Dommermuth}}\ and\ \bibinfo {author} {\bibfnamefont {D.~K.~P.}\ \bibnamefont {Yue}},\ }\bibfield  {title} {\enquote {\bibinfo {title} {A high-order spectral method for the study of nonlinear gravity waves},}\ }\href {\doibase 10.1017/S002211208700288X} {\bibfield  {journal} {\bibinfo  {journal} {Journal of Fluid Mechanics}\ }\textbf {\bibinfo {volume} {184}},\ \bibinfo {pages} {267–288} (\bibinfo {year} {1987})}\BibitemShut {NoStop}%
\bibitem [{\citenamefont {West}\ \emph {et~al.}(1987)\citenamefont {West}, \citenamefont {Brueckner}, \citenamefont {Janda}, \citenamefont {Milder},\ and\ \citenamefont {Milton}}]{West1987}%
  \BibitemOpen
  \bibfield  {author} {\bibinfo {author} {\bibfnamefont {B.~J.}\ \bibnamefont {West}}, \bibinfo {author} {\bibfnamefont {K.~A.}\ \bibnamefont {Brueckner}}, \bibinfo {author} {\bibfnamefont {R.~S.}\ \bibnamefont {Janda}}, \bibinfo {author} {\bibfnamefont {D.~M.}\ \bibnamefont {Milder}}, \ and\ \bibinfo {author} {\bibfnamefont {R.~L.}\ \bibnamefont {Milton}},\ }\bibfield  {title} {\enquote {\bibinfo {title} {A new numerical method for surface hydrodynamics},}\ }\href {\doibase 10.1029/JC092iC11p11803} {\bibfield  {journal} {\bibinfo  {journal} {Journal of Geophysical Research: Oceans}\ }\textbf {\bibinfo {volume} {92}},\ \bibinfo {pages} {11803--11824} (\bibinfo {year} {1987})}\BibitemShut {NoStop}%
\bibitem [{\citenamefont {Wu}(2004)}]{Wu2004}%
  \BibitemOpen
  \bibfield  {author} {\bibinfo {author} {\bibfnamefont {G.}~\bibnamefont {Wu}},\ }\emph {\bibinfo {title} {Direct simulation and deterministic prediction of large-scale nonlinear ocean wave-field}},\ \href@noop {} {Ph.D. thesis},\ \bibinfo  {school} {Massachusetts Institute of Technology} (\bibinfo {year} {2004})\BibitemShut {NoStop}%
\bibitem [{\citenamefont {Ducrozet}\ \emph {et~al.}(2007)\citenamefont {Ducrozet}, \citenamefont {Bonnefoy}, \citenamefont {Le~Touz{\'e}},\ and\ \citenamefont {Ferrant}}]{Ducrozet2007}%
  \BibitemOpen
  \bibfield  {author} {\bibinfo {author} {\bibfnamefont {G.}~\bibnamefont {Ducrozet}}, \bibinfo {author} {\bibfnamefont {F.}~\bibnamefont {Bonnefoy}}, \bibinfo {author} {\bibfnamefont {D.}~\bibnamefont {Le~Touz{\'e}}}, \ and\ \bibinfo {author} {\bibfnamefont {P.}~\bibnamefont {Ferrant}},\ }\bibfield  {title} {\enquote {\bibinfo {title} {3-d hos simulations of extreme waves in open seas},}\ }\href@noop {} {\bibfield  {journal} {\bibinfo  {journal} {Natural Hazards and Earth System Sciences}\ }\textbf {\bibinfo {volume} {7}},\ \bibinfo {pages} {109--122} (\bibinfo {year} {2007})}\BibitemShut {NoStop}%
\bibitem [{\citenamefont {Ducrozet}\ \emph {et~al.}(2016)\citenamefont {Ducrozet}, \citenamefont {Bonnefoy}, \citenamefont {Le~Touz{\'e}},\ and\ \citenamefont {Ferrant}}]{Ducrozet2016}%
  \BibitemOpen
  \bibfield  {author} {\bibinfo {author} {\bibfnamefont {G.}~\bibnamefont {Ducrozet}}, \bibinfo {author} {\bibfnamefont {F.}~\bibnamefont {Bonnefoy}}, \bibinfo {author} {\bibfnamefont {D.}~\bibnamefont {Le~Touz{\'e}}}, \ and\ \bibinfo {author} {\bibfnamefont {P.}~\bibnamefont {Ferrant}},\ }\bibfield  {title} {\enquote {\bibinfo {title} {Hos-ocean: Open-source solver for nonlinear waves in open ocean based on high-order spectral method},}\ }\href@noop {} {\bibfield  {journal} {\bibinfo  {journal} {Computer Physics Communications}\ }\textbf {\bibinfo {volume} {203}},\ \bibinfo {pages} {245--254} (\bibinfo {year} {2016})}\BibitemShut {NoStop}%
\bibitem [{\citenamefont {Bonnefoy}\ \emph {et~al.}(2010)\citenamefont {Bonnefoy}, \citenamefont {Ducrozet}, \citenamefont {Le~Touz{\'e}},\ and\ \citenamefont {Ferrant}}]{Bonnefoy2010}%
  \BibitemOpen
  \bibfield  {author} {\bibinfo {author} {\bibfnamefont {F.}~\bibnamefont {Bonnefoy}}, \bibinfo {author} {\bibfnamefont {G.}~\bibnamefont {Ducrozet}}, \bibinfo {author} {\bibfnamefont {D.}~\bibnamefont {Le~Touz{\'e}}}, \ and\ \bibinfo {author} {\bibfnamefont {P.}~\bibnamefont {Ferrant}},\ }\bibfield  {title} {\enquote {\bibinfo {title} {Time domain simulation of nonlinear water waves using spectral methods},}\ }in\ \href@noop {} {\emph {\bibinfo {booktitle} {Advances in numerical simulation of nonlinear water waves}}}\ (\bibinfo  {publisher} {World Scientific},\ \bibinfo {year} {2010})\ pp.\ \bibinfo {pages} {129--164}\BibitemShut {NoStop}%
\bibitem [{\citenamefont {Klein}\ \emph {et~al.}(2020)\citenamefont {Klein}, \citenamefont {Dudek}, \citenamefont {Clauss}, \citenamefont {Ehlers}, \citenamefont {Behrendt}, \citenamefont {Hoffmann},\ and\ \citenamefont {Onorato}}]{Klein2020}%
  \BibitemOpen
  \bibfield  {author} {\bibinfo {author} {\bibfnamefont {M.}~\bibnamefont {Klein}}, \bibinfo {author} {\bibfnamefont {M.}~\bibnamefont {Dudek}}, \bibinfo {author} {\bibfnamefont {G.~F.}\ \bibnamefont {Clauss}}, \bibinfo {author} {\bibfnamefont {S.}~\bibnamefont {Ehlers}}, \bibinfo {author} {\bibfnamefont {J.}~\bibnamefont {Behrendt}}, \bibinfo {author} {\bibfnamefont {N.}~\bibnamefont {Hoffmann}}, \ and\ \bibinfo {author} {\bibfnamefont {M.}~\bibnamefont {Onorato}},\ }\bibfield  {title} {\enquote {\bibinfo {title} {On the deterministic prediction of water waves},}\ }\href {\doibase 10.3390/fluids5010009} {\bibfield  {journal} {\bibinfo  {journal} {Fluids}\ }\textbf {\bibinfo {volume} {5}} (\bibinfo {year} {2020}),\ 10.3390/fluids5010009}\BibitemShut {NoStop}%
\bibitem [{\citenamefont {Ducrozet}\ \emph {et~al.}(2012)\citenamefont {Ducrozet}, \citenamefont {Bingham}, \citenamefont {Engsig-Karup}, \citenamefont {Bonnefoy},\ and\ \citenamefont {Ferrant}}]{Ducrozet2012}%
  \BibitemOpen
  \bibfield  {author} {\bibinfo {author} {\bibfnamefont {G.}~\bibnamefont {Ducrozet}}, \bibinfo {author} {\bibfnamefont {H.~B.}\ \bibnamefont {Bingham}}, \bibinfo {author} {\bibfnamefont {A.~P.}\ \bibnamefont {Engsig-Karup}}, \bibinfo {author} {\bibfnamefont {F.}~\bibnamefont {Bonnefoy}}, \ and\ \bibinfo {author} {\bibfnamefont {P.}~\bibnamefont {Ferrant}},\ }\bibfield  {title} {\enquote {\bibinfo {title} {A comparative study of two fast nonlinear free-surface water wave models},}\ }\href {\doibase https://doi.org/10.1002/fld.2672} {\bibfield  {journal} {\bibinfo  {journal} {International Journal for Numerical Methods in Fluids}\ }\textbf {\bibinfo {volume} {69}},\ \bibinfo {pages} {1818--1834} (\bibinfo {year} {2012})}\BibitemShut {NoStop}%
\bibitem [{\citenamefont {Köllisch}\ \emph {et~al.}(2018)\citenamefont {Köllisch}, \citenamefont {Behrendt}, \citenamefont {Klein},\ and\ \citenamefont {Hoffmann}}]{Koellisch2018}%
  \BibitemOpen
  \bibfield  {author} {\bibinfo {author} {\bibfnamefont {N.}~\bibnamefont {Köllisch}}, \bibinfo {author} {\bibfnamefont {J.}~\bibnamefont {Behrendt}}, \bibinfo {author} {\bibfnamefont {M.}~\bibnamefont {Klein}}, \ and\ \bibinfo {author} {\bibfnamefont {N.}~\bibnamefont {Hoffmann}},\ }\bibfield  {title} {\enquote {\bibinfo {title} {Nonlinear real time prediction of ocean surface waves},}\ }\href {\doibase 10.1016/j.oceaneng.2018.03.048} {\bibfield  {journal} {\bibinfo  {journal} {Ocean Engineering}\ }\textbf {\bibinfo {volume} {157}},\ \bibinfo {pages} {387--400} (\bibinfo {year} {2018})}\BibitemShut {NoStop}%
\bibitem [{\citenamefont {Hlophe}(2023)}]{Hlophe2023}%
  \BibitemOpen
  \bibfield  {author} {\bibinfo {author} {\bibfnamefont {T.}~\bibnamefont {Hlophe}},\ }\emph {\bibinfo {title} {Efficient real-time deterministic wave prediction for marine renewable energy}},\ \href {\doibase 10.26182/t850-y895} {Ph.D. thesis},\ \bibinfo  {school} {The University of Western Australia} (\bibinfo {year} {2023})\BibitemShut {NoStop}%
\bibitem [{\citenamefont {Yoon}, \citenamefont {Kim},\ and\ \citenamefont {Choi}(2016)}]{Yoon2016}%
  \BibitemOpen
  \bibfield  {author} {\bibinfo {author} {\bibfnamefont {S.}~\bibnamefont {Yoon}}, \bibinfo {author} {\bibfnamefont {J.}~\bibnamefont {Kim}}, \ and\ \bibinfo {author} {\bibfnamefont {W.}~\bibnamefont {Choi}},\ }\bibfield  {title} {\enquote {\bibinfo {title} {An explicit data assimilation scheme for a nonlinear wave prediction model based on a pseudo-spectral method},}\ }\href {\doibase 10.1109/JOE.2015.2406471} {\bibfield  {journal} {\bibinfo  {journal} {IEEE Journal of Oceanic Engineering}\ }\textbf {\bibinfo {volume} {41}},\ \bibinfo {pages} {112--122} (\bibinfo {year} {2016})}\BibitemShut {NoStop}%
\bibitem [{\citenamefont {Wang}\ and\ \citenamefont {Pan}(2021)}]{Wang2021}%
  \BibitemOpen
  \bibfield  {author} {\bibinfo {author} {\bibfnamefont {G.}~\bibnamefont {Wang}}\ and\ \bibinfo {author} {\bibfnamefont {Y.}~\bibnamefont {Pan}},\ }\bibfield  {title} {\enquote {\bibinfo {title} {Phase-resolved ocean wave forecast with ensemble-based data assimilation},}\ }\href {\doibase 10.1017/jfm.2021.340} {\bibfield  {journal} {\bibinfo  {journal} {Journal of Fluid Mechanics}\ }\textbf {\bibinfo {volume} {918}},\ \bibinfo {pages} {A19} (\bibinfo {year} {2021})}\BibitemShut {NoStop}%
\bibitem [{\citenamefont {Blondel}, \citenamefont {Bonnefoy},\ and\ \citenamefont {Ferrant}(2010)}]{Blondel2010}%
  \BibitemOpen
  \bibfield  {author} {\bibinfo {author} {\bibfnamefont {E.}~\bibnamefont {Blondel}}, \bibinfo {author} {\bibfnamefont {F.}~\bibnamefont {Bonnefoy}}, \ and\ \bibinfo {author} {\bibfnamefont {P.}~\bibnamefont {Ferrant}},\ }\bibfield  {title} {\enquote {\bibinfo {title} {Deterministic non-linear wave prediction using probe data},}\ }\href {\doibase 10.1016/j.oceaneng.2010.03.002} {\bibfield  {journal} {\bibinfo  {journal} {Ocean Engineering}\ }\textbf {\bibinfo {volume} {37}},\ \bibinfo {pages} {913--926} (\bibinfo {year} {2010})}\BibitemShut {NoStop}%
\bibitem [{\citenamefont {Aragh}\ and\ \citenamefont {Nwogu}(2008)}]{Aragh2008}%
  \BibitemOpen
  \bibfield  {author} {\bibinfo {author} {\bibfnamefont {S.}~\bibnamefont {Aragh}}\ and\ \bibinfo {author} {\bibfnamefont {O.}~\bibnamefont {Nwogu}},\ }\bibfield  {title} {\enquote {\bibinfo {title} {Variation assimilating of synthetic radar data into a pseudo-spectral wave model},}\ }\href {\doibase 10.2112/1551-5036-52.sp1.235} {\bibfield  {journal} {\bibinfo  {journal} {Journal of Coastal Research}\ ,\ \bibinfo {pages} {235--244}} (\bibinfo {year} {2008})}\BibitemShut {NoStop}%
\bibitem [{\citenamefont {Qi}\ \emph {et~al.}(2018)\citenamefont {Qi}, \citenamefont {Wu}, \citenamefont {Liu}, \citenamefont {Kim},\ and\ \citenamefont {Yue}}]{Qi2018}%
  \BibitemOpen
  \bibfield  {author} {\bibinfo {author} {\bibfnamefont {Y.}~\bibnamefont {Qi}}, \bibinfo {author} {\bibfnamefont {G.}~\bibnamefont {Wu}}, \bibinfo {author} {\bibfnamefont {Y.}~\bibnamefont {Liu}}, \bibinfo {author} {\bibfnamefont {M.-H.}\ \bibnamefont {Kim}}, \ and\ \bibinfo {author} {\bibfnamefont {D.~K.~P.}\ \bibnamefont {Yue}},\ }\bibfield  {title} {\enquote {\bibinfo {title} {Nonlinear phase-resolved reconstruction of irregular water waves},}\ }\href {\doibase 10.1017/jfm.2017.904} {\bibfield  {journal} {\bibinfo  {journal} {Journal of Fluid Mechanics}\ }\textbf {\bibinfo {volume} {838}},\ \bibinfo {pages} {544–572} (\bibinfo {year} {2018})}\BibitemShut {NoStop}%
\bibitem [{\citenamefont {Fujimoto}\ and\ \citenamefont {Waseda}(2020)}]{Fujimoto2020}%
  \BibitemOpen
  \bibfield  {author} {\bibinfo {author} {\bibfnamefont {W.}~\bibnamefont {Fujimoto}}\ and\ \bibinfo {author} {\bibfnamefont {T.}~\bibnamefont {Waseda}},\ }\bibfield  {title} {\enquote {\bibinfo {title} {Ensemble-based variational method for nonlinear inversion of surface gravity waves},}\ }\href {\doibase 10.1175/JTECH-D-19-0072.1} {\bibfield  {journal} {\bibinfo  {journal} {Journal of Atmospheric and Oceanic Technology}\ }\textbf {\bibinfo {volume} {37}},\ \bibinfo {pages} {17 -- 31} (\bibinfo {year} {2020})}\BibitemShut {NoStop}%
\bibitem [{\citenamefont {Wu}, \citenamefont {Hao},\ and\ \citenamefont {Shen}(2022)}]{Wu2022}%
  \BibitemOpen
  \bibfield  {author} {\bibinfo {author} {\bibfnamefont {J.}~\bibnamefont {Wu}}, \bibinfo {author} {\bibfnamefont {X.}~\bibnamefont {Hao}}, \ and\ \bibinfo {author} {\bibfnamefont {L.}~\bibnamefont {Shen}},\ }\bibfield  {title} {\enquote {\bibinfo {title} {An improved adjoint-based ocean wave reconstruction and prediction method},}\ }\href {\doibase 10.1017/flo.2021.19} {\bibfield  {journal} {\bibinfo  {journal} {Flow}\ }\textbf {\bibinfo {volume} {2}},\ \bibinfo {pages} {E2} (\bibinfo {year} {2022})}\BibitemShut {NoStop}%
\bibitem [{Des(2021)}]{Desmars2021}%
  \BibitemOpen
  \href {\doibase 10.1115/OMAE2021-62409} {\emph {\bibinfo {title} {Reconstruction of Ocean Surfaces From Randomly Distributed Measurements Using a Grid-Based Method}}},\ \bibinfo {series} {International Conference on Offshore Mechanics and Arctic Engineering}, Vol.\ \bibinfo {volume} {Volume 6: Ocean Engineering}\ (\bibinfo {year} {2021})\BibitemShut {NoStop}%
\bibitem [{\citenamefont {Desmars}\ \emph {et~al.}(2023)\citenamefont {Desmars}, \citenamefont {Hartmann}, \citenamefont {Behrendt}, \citenamefont {Hoffmann},\ and\ \citenamefont {Klein}}]{Desmars2023}%
  \BibitemOpen
  \bibfield  {author} {\bibinfo {author} {\bibfnamefont {N.}~\bibnamefont {Desmars}}, \bibinfo {author} {\bibfnamefont {M.}~\bibnamefont {Hartmann}}, \bibinfo {author} {\bibfnamefont {J.}~\bibnamefont {Behrendt}}, \bibinfo {author} {\bibfnamefont {N.}~\bibnamefont {Hoffmann}}, \ and\ \bibinfo {author} {\bibfnamefont {M.}~\bibnamefont {Klein}},\ }\bibfield  {title} {\enquote {\bibinfo {title} {Nonlinear deterministic reconstruction and prediction of remotely measured ocean surface waves},}\ }\href {\doibase 10.1017/jfm.2023.841} {\bibfield  {journal} {\bibinfo  {journal} {Journal of Fluid Mechanics}\ }\textbf {\bibinfo {volume} {975}},\ \bibinfo {pages} {A8} (\bibinfo {year} {2023})}\BibitemShut {NoStop}%
\bibitem [{\citenamefont {Desmars}(2020)}]{Desmars2020}%
  \BibitemOpen
  \bibfield  {author} {\bibinfo {author} {\bibfnamefont {N.}~\bibnamefont {Desmars}},\ }\emph {\bibinfo {title} {{Real-time reconstruction and prediction of ocean wave fields from remote optical measurements}}},\ \href {https://theses.hal.science/tel-03185222} {\bibinfo {type} {Theses}},\ \bibinfo  {school} {{{\'E}cole centrale de Nantes}} (\bibinfo {year} {2020})\BibitemShut {NoStop}%
\bibitem [{\citenamefont {Ehlers}\ \emph {et~al.}(2023)\citenamefont {Ehlers}, \citenamefont {Klein}, \citenamefont {Heinlein}, \citenamefont {Wedler}, \citenamefont {Desmars}, \citenamefont {Hoffmann},\ and\ \citenamefont {Stender}}]{Ehlers2023}%
  \BibitemOpen
  \bibfield  {author} {\bibinfo {author} {\bibfnamefont {S.}~\bibnamefont {Ehlers}}, \bibinfo {author} {\bibfnamefont {M.}~\bibnamefont {Klein}}, \bibinfo {author} {\bibfnamefont {A.}~\bibnamefont {Heinlein}}, \bibinfo {author} {\bibfnamefont {M.}~\bibnamefont {Wedler}}, \bibinfo {author} {\bibfnamefont {N.}~\bibnamefont {Desmars}}, \bibinfo {author} {\bibfnamefont {N.}~\bibnamefont {Hoffmann}}, \ and\ \bibinfo {author} {\bibfnamefont {M.}~\bibnamefont {Stender}},\ }\bibfield  {title} {\enquote {\bibinfo {title} {Machine learning for phase-resolved reconstruction of nonlinear ocean wave surface elevations from sparse remote sensing data},}\ }\href {\doibase 10.1016/j.oceaneng.2023.116059} {\bibfield  {journal} {\bibinfo  {journal} {Ocean Engineering}\ }\textbf {\bibinfo {volume} {288}},\ \bibinfo {pages} {116059} (\bibinfo {year} {2023})}\BibitemShut {NoStop}%
\bibitem [{\citenamefont {Li}\ \emph {et~al.}(2020)\citenamefont {Li}, \citenamefont {Kovachki}, \citenamefont {Azizzadenesheli}, \citenamefont {Liu}, \citenamefont {Bhattacharya}, \citenamefont {Stuart},\ and\ \citenamefont {Anandkumar}}]{Li2020}%
  \BibitemOpen
  \bibfield  {author} {\bibinfo {author} {\bibfnamefont {Z.}~\bibnamefont {Li}}, \bibinfo {author} {\bibfnamefont {N.}~\bibnamefont {Kovachki}}, \bibinfo {author} {\bibfnamefont {K.}~\bibnamefont {Azizzadenesheli}}, \bibinfo {author} {\bibfnamefont {B.}~\bibnamefont {Liu}}, \bibinfo {author} {\bibfnamefont {K.}~\bibnamefont {Bhattacharya}}, \bibinfo {author} {\bibfnamefont {A.}~\bibnamefont {Stuart}}, \ and\ \bibinfo {author} {\bibfnamefont {A.}~\bibnamefont {Anandkumar}},\ }\bibfield  {title} {\enquote {\bibinfo {title} {Fourier neural operator for parametric partial differential equations},}\ }\href@noop {} {\bibfield  {journal} {\bibinfo  {journal} {arXiv preprint arXiv:2010.08895}\ } (\bibinfo {year} {2020})}\BibitemShut {NoStop}%
\bibitem [{\citenamefont {Ehlers}\ \emph {et~al.}(2025)\citenamefont {Ehlers}, \citenamefont {Hoffmann}, \citenamefont {Tang}, \citenamefont {Callaghan}, \citenamefont {Cao}, \citenamefont {Padilla}, \citenamefont {Fang},\ and\ \citenamefont {Stender}}]{Ehlers2025}%
  \BibitemOpen
  \bibfield  {author} {\bibinfo {author} {\bibfnamefont {S.}~\bibnamefont {Ehlers}}, \bibinfo {author} {\bibfnamefont {N.}~\bibnamefont {Hoffmann}}, \bibinfo {author} {\bibfnamefont {T.}~\bibnamefont {Tang}}, \bibinfo {author} {\bibfnamefont {A.~H.}\ \bibnamefont {Callaghan}}, \bibinfo {author} {\bibfnamefont {R.}~\bibnamefont {Cao}}, \bibinfo {author} {\bibfnamefont {E.~M.}\ \bibnamefont {Padilla}}, \bibinfo {author} {\bibfnamefont {Y.}~\bibnamefont {Fang}}, \ and\ \bibinfo {author} {\bibfnamefont {M.}~\bibnamefont {Stender}},\ }\bibfield  {title} {\enquote {\bibinfo {title} {Physics-informed neural networks for phase-resolved data assimilation and prediction of nonlinear ocean waves},}\ }\href {https://arxiv.org/abs/2501.08430} {\bibfield  {journal} {\bibinfo  {journal} {Physical Review Fluids}\ } (\bibinfo {year} {2025})}\BibitemShut {NoStop}%
\bibitem [{\citenamefont {Raissi}, \citenamefont {Perdikaris},\ and\ \citenamefont {Karniadakis}(2019)}]{Raissi2019}%
  \BibitemOpen
  \bibfield  {author} {\bibinfo {author} {\bibfnamefont {M.}~\bibnamefont {Raissi}}, \bibinfo {author} {\bibfnamefont {P.}~\bibnamefont {Perdikaris}}, \ and\ \bibinfo {author} {\bibfnamefont {G.}~\bibnamefont {Karniadakis}},\ }\bibfield  {title} {\enquote {\bibinfo {title} {Physics-informed neural networks: A deep learning framework for solving forward and inverse problems involving nonlinear partial differential equations},}\ }\href {\doibase 10.1016/j.jcp.2018.10.045} {\bibfield  {journal} {\bibinfo  {journal} {Journal of Computational Physics}\ }\textbf {\bibinfo {volume} {378}},\ \bibinfo {pages} {686--707} (\bibinfo {year} {2019})}\BibitemShut {NoStop}%
\bibitem [{\citenamefont {Law}\ \emph {et~al.}(2020)\citenamefont {Law}, \citenamefont {Santo}, \citenamefont {Lim},\ and\ \citenamefont {Chan}}]{Law2020}%
  \BibitemOpen
  \bibfield  {author} {\bibinfo {author} {\bibfnamefont {Y.}~\bibnamefont {Law}}, \bibinfo {author} {\bibfnamefont {H.}~\bibnamefont {Santo}}, \bibinfo {author} {\bibfnamefont {K.}~\bibnamefont {Lim}}, \ and\ \bibinfo {author} {\bibfnamefont {E.}~\bibnamefont {Chan}},\ }\bibfield  {title} {\enquote {\bibinfo {title} {Deterministic wave prediction for unidirectional sea-states in real-time using artificial neural network},}\ }\href {\doibase 10.1016/j.oceaneng.2019.106722} {\bibfield  {journal} {\bibinfo  {journal} {Ocean Engineering}\ }\textbf {\bibinfo {volume} {195}},\ \bibinfo {pages} {106722} (\bibinfo {year} {2020})}\BibitemShut {NoStop}%
\bibitem [{\citenamefont {Zhang}\ \emph {et~al.}(2022)\citenamefont {Zhang}, \citenamefont {Zhao}, \citenamefont {Jin},\ and\ \citenamefont {Greaves}}]{Zhang2022}%
  \BibitemOpen
  \bibfield  {author} {\bibinfo {author} {\bibfnamefont {J.}~\bibnamefont {Zhang}}, \bibinfo {author} {\bibfnamefont {X.}~\bibnamefont {Zhao}}, \bibinfo {author} {\bibfnamefont {S.}~\bibnamefont {Jin}}, \ and\ \bibinfo {author} {\bibfnamefont {D.}~\bibnamefont {Greaves}},\ }\bibfield  {title} {\enquote {\bibinfo {title} {Phase-resolved real-time ocean wave prediction with quantified uncertainty based on variational bayesian machine learning},}\ }\href {\doibase 10.1016/j.apenergy.2022.119711} {\bibfield  {journal} {\bibinfo  {journal} {Applied Energy}\ }\textbf {\bibinfo {volume} {324}},\ \bibinfo {pages} {119711} (\bibinfo {year} {2022})}\BibitemShut {NoStop}%
\bibitem [{\citenamefont {{Le Quang}}, \citenamefont {Dao},\ and\ \citenamefont {Lu}(2023)}]{LeQuang2023}%
  \BibitemOpen
  \bibfield  {author} {\bibinfo {author} {\bibfnamefont {T.}~\bibnamefont {{Le Quang}}}, \bibinfo {author} {\bibfnamefont {M.~H.}\ \bibnamefont {Dao}}, \ and\ \bibinfo {author} {\bibfnamefont {X.}~\bibnamefont {Lu}},\ }\bibfield  {title} {\enquote {\bibinfo {title} {Prediction of near-field uni-directional and multi-directional random waves from far-field measurements with artificial neural networks},}\ }\href {\doibase https://doi.org/10.1016/j.oceaneng.2023.114307} {\bibfield  {journal} {\bibinfo  {journal} {Ocean Engineering}\ }\textbf {\bibinfo {volume} {278}},\ \bibinfo {pages} {114307} (\bibinfo {year} {2023})}\BibitemShut {NoStop}%
\bibitem [{\citenamefont {Liu}\ \emph {et~al.}(2024)\citenamefont {Liu}, \citenamefont {Zhang}, \citenamefont {Chen}, \citenamefont {Dong}, \citenamefont {Guo}, \citenamefont {Tian}, \citenamefont {Lu},\ and\ \citenamefont {Peng}}]{Liu2024}%
  \BibitemOpen
  \bibfield  {author} {\bibinfo {author} {\bibfnamefont {Y.}~\bibnamefont {Liu}}, \bibinfo {author} {\bibfnamefont {X.}~\bibnamefont {Zhang}}, \bibinfo {author} {\bibfnamefont {G.}~\bibnamefont {Chen}}, \bibinfo {author} {\bibfnamefont {Q.}~\bibnamefont {Dong}}, \bibinfo {author} {\bibfnamefont {X.}~\bibnamefont {Guo}}, \bibinfo {author} {\bibfnamefont {X.}~\bibnamefont {Tian}}, \bibinfo {author} {\bibfnamefont {W.}~\bibnamefont {Lu}}, \ and\ \bibinfo {author} {\bibfnamefont {T.}~\bibnamefont {Peng}},\ }\bibfield  {title} {\enquote {\bibinfo {title} {Deterministic wave prediction model for irregular long-crested waves with recurrent neural network},}\ }\href {\doibase https://doi.org/10.1016/j.joes.2022.08.002} {\bibfield  {journal} {\bibinfo  {journal} {Journal of Ocean Engineering and Science}\ }\textbf {\bibinfo {volume} {9}},\ \bibinfo {pages} {251--263} (\bibinfo {year} {2024})}\BibitemShut {NoStop}%
\bibitem [{\citenamefont {Mohaghegh}, \citenamefont {Murthy},\ and\ \citenamefont {Alam}(2021)}]{Mohaghegh2021}%
  \BibitemOpen
  \bibfield  {author} {\bibinfo {author} {\bibfnamefont {F.}~\bibnamefont {Mohaghegh}}, \bibinfo {author} {\bibfnamefont {J.}~\bibnamefont {Murthy}}, \ and\ \bibinfo {author} {\bibfnamefont {M.-R.}\ \bibnamefont {Alam}},\ }\bibfield  {title} {\enquote {\bibinfo {title} {Rapid phase-resolved prediction of nonlinear dispersive waves using machine learning},}\ }\href {\doibase 10.1016/j.apor.2021.102920} {\bibfield  {journal} {\bibinfo  {journal} {Applied Ocean Research}\ }\textbf {\bibinfo {volume} {117}},\ \bibinfo {pages} {102920} (\bibinfo {year} {2021})}\BibitemShut {NoStop}%
\bibitem [{\citenamefont {Wedler}\ \emph {et~al.}(2023)\citenamefont {Wedler}, \citenamefont {Stender}, \citenamefont {Klein},\ and\ \citenamefont {Hoffmann}}]{Wedler2023}%
  \BibitemOpen
  \bibfield  {author} {\bibinfo {author} {\bibfnamefont {M.}~\bibnamefont {Wedler}}, \bibinfo {author} {\bibfnamefont {M.}~\bibnamefont {Stender}}, \bibinfo {author} {\bibfnamefont {M.}~\bibnamefont {Klein}}, \ and\ \bibinfo {author} {\bibfnamefont {N.}~\bibnamefont {Hoffmann}},\ }\bibfield  {title} {\enquote {\bibinfo {title} {Machine learning simulation of one-dimensional deterministic water wave propagation},}\ }\href {\doibase 10.1016/j.oceaneng.2023.115222} {\bibfield  {journal} {\bibinfo  {journal} {Ocean Engineering}\ }\textbf {\bibinfo {volume} {284}},\ \bibinfo {pages} {115222} (\bibinfo {year} {2023})}\BibitemShut {NoStop}%
\bibitem [{\citenamefont {Li}\ \emph {et~al.}(2024)\citenamefont {Li}, \citenamefont {Zheng}, \citenamefont {Kovachki}, \citenamefont {Jin}, \citenamefont {Chen}, \citenamefont {Liu}, \citenamefont {Azizzadenesheli},\ and\ \citenamefont {Anandkumar}}]{Li2024}%
  \BibitemOpen
  \bibfield  {author} {\bibinfo {author} {\bibfnamefont {Z.}~\bibnamefont {Li}}, \bibinfo {author} {\bibfnamefont {H.}~\bibnamefont {Zheng}}, \bibinfo {author} {\bibfnamefont {N.}~\bibnamefont {Kovachki}}, \bibinfo {author} {\bibfnamefont {D.}~\bibnamefont {Jin}}, \bibinfo {author} {\bibfnamefont {H.}~\bibnamefont {Chen}}, \bibinfo {author} {\bibfnamefont {B.}~\bibnamefont {Liu}}, \bibinfo {author} {\bibfnamefont {K.}~\bibnamefont {Azizzadenesheli}}, \ and\ \bibinfo {author} {\bibfnamefont {A.}~\bibnamefont {Anandkumar}},\ }\bibfield  {title} {\enquote {\bibinfo {title} {Physics-informed neural operator for learning partial differential equations},}\ }\href {\doibase 10.1145/3648506} {\bibfield  {journal} {\bibinfo  {journal} {ACM / IMS J. Data Sci.}\ }\textbf {\bibinfo {volume} {1}} (\bibinfo {year} {2024}),\ 10.1145/3648506}\BibitemShut {NoStop}%
\bibitem [{\citenamefont {Rosofsky}, \citenamefont {Al~Majed},\ and\ \citenamefont {Huerta}(2023)}]{Rosofsky2023}%
  \BibitemOpen
  \bibfield  {author} {\bibinfo {author} {\bibfnamefont {S.~G.}\ \bibnamefont {Rosofsky}}, \bibinfo {author} {\bibfnamefont {H.}~\bibnamefont {Al~Majed}}, \ and\ \bibinfo {author} {\bibfnamefont {E.~A.}\ \bibnamefont {Huerta}},\ }\bibfield  {title} {\enquote {\bibinfo {title} {Applications of physics informed neural operators},}\ }\href {\doibase 10.1088/2632-2153/acd168} {\bibfield  {journal} {\bibinfo  {journal} {Machine Learning: Science and Technology}\ }\textbf {\bibinfo {volume} {4}},\ \bibinfo {pages} {025022} (\bibinfo {year} {2023})}\BibitemShut {NoStop}%
\bibitem [{\citenamefont {Zhao}\ \emph {et~al.}(2024)\citenamefont {Zhao}, \citenamefont {Li}, \citenamefont {Fan}, \citenamefont {Wang}, \citenamefont {Yang},\ and\ \citenamefont {Wang}}]{Zhao2024}%
  \BibitemOpen
  \bibfield  {author} {\bibinfo {author} {\bibfnamefont {S.}~\bibnamefont {Zhao}}, \bibinfo {author} {\bibfnamefont {Z.}~\bibnamefont {Li}}, \bibinfo {author} {\bibfnamefont {B.}~\bibnamefont {Fan}}, \bibinfo {author} {\bibfnamefont {Y.}~\bibnamefont {Wang}}, \bibinfo {author} {\bibfnamefont {H.}~\bibnamefont {Yang}}, \ and\ \bibinfo {author} {\bibfnamefont {J.}~\bibnamefont {Wang}},\ }\href {https://arxiv.org/abs/2411.04502} {\enquote {\bibinfo {title} {Lesnets (large-eddy simulation nets): Physics-informed neural operator for large-eddy simulation of turbulence},}\ } (\bibinfo {year} {2024}),\ \Eprint {http://arxiv.org/abs/2411.04502} {arXiv:2411.04502 [physics.flu-dyn]} \BibitemShut {NoStop}%
\bibitem [{\citenamefont {Konuk}\ and\ \citenamefont {Shragge}(2021)}]{Konuk2021}%
  \BibitemOpen
  \bibfield  {author} {\bibinfo {author} {\bibfnamefont {T.}~\bibnamefont {Konuk}}\ and\ \bibinfo {author} {\bibfnamefont {J.}~\bibnamefont {Shragge}},\ }\bibfield  {title} {\enquote {\bibinfo {title} {Physics-guided deep learning using fourier neural operators for solving the acoustic vti wave equation},}\ }in\ \href@noop {} {\emph {\bibinfo {booktitle} {82nd EAGE annual conference \& exhibition}}},\ \bibinfo {series and number} {\bibinfo {number} {1}}\ (\bibinfo {organization} {European Association of Geoscientists \& Engineers},\ \bibinfo {year} {2021})\ pp.\ \bibinfo {pages} {1--5}\BibitemShut {NoStop}%
\bibitem [{\citenamefont {Wang}\ \emph {et~al.}(2024)\citenamefont {Wang}, \citenamefont {Jiang}, \citenamefont {Yan}, \citenamefont {He}, \citenamefont {Feng}, \citenamefont {Pan},\ and\ \citenamefont {Luo}}]{Wang2024}%
  \BibitemOpen
  \bibfield  {author} {\bibinfo {author} {\bibfnamefont {Q.}~\bibnamefont {Wang}}, \bibinfo {author} {\bibfnamefont {L.}~\bibnamefont {Jiang}}, \bibinfo {author} {\bibfnamefont {L.}~\bibnamefont {Yan}}, \bibinfo {author} {\bibfnamefont {X.}~\bibnamefont {He}}, \bibinfo {author} {\bibfnamefont {J.}~\bibnamefont {Feng}}, \bibinfo {author} {\bibfnamefont {W.}~\bibnamefont {Pan}}, \ and\ \bibinfo {author} {\bibfnamefont {B.}~\bibnamefont {Luo}},\ }\bibfield  {title} {\enquote {\bibinfo {title} {Chaotic time series prediction based on physics-informed neural operator},}\ }\href {\doibase https://doi.org/10.1016/j.chaos.2024.115326} {\bibfield  {journal} {\bibinfo  {journal} {Chaos, Solitons \& Fractals}\ }\textbf {\bibinfo {volume} {186}},\ \bibinfo {pages} {115326} (\bibinfo {year} {2024})}\BibitemShut {NoStop}%
\bibitem [{\citenamefont {Chen}\ and\ \citenamefont {Chen}(1995)}]{Chen1995}%
  \BibitemOpen
  \bibfield  {author} {\bibinfo {author} {\bibfnamefont {T.}~\bibnamefont {Chen}}\ and\ \bibinfo {author} {\bibfnamefont {H.}~\bibnamefont {Chen}},\ }\bibfield  {title} {\enquote {\bibinfo {title} {Universal approximation to nonlinear operators by neural networks with arbitrary activation functions and its application to dynamical systems},}\ }\href@noop {} {\bibfield  {journal} {\bibinfo  {journal} {IEEE transactions on neural networks}\ }\textbf {\bibinfo {volume} {6}},\ \bibinfo {pages} {911--917} (\bibinfo {year} {1995})}\BibitemShut {NoStop}%
\bibitem [{\citenamefont {Chiang}, \citenamefont {Stiassnie},\ and\ \citenamefont {Yue}(2005)}]{Chiang2005}%
  \BibitemOpen
  \bibfield  {author} {\bibinfo {author} {\bibfnamefont {C.~M.}\ \bibnamefont {Chiang}}, \bibinfo {author} {\bibfnamefont {M.}~\bibnamefont {Stiassnie}}, \ and\ \bibinfo {author} {\bibfnamefont {D.~K.-P.}\ \bibnamefont {Yue}},\ }\href {\doibase 10.1142/5566} {\emph {\bibinfo {title} {Theory and Applications of Ocean Surface Waves}}}\ (\bibinfo  {publisher} {World Scientific},\ \bibinfo {year} {2005})\ \Eprint {http://arxiv.org/abs/https://www.worldscientific.com/doi/pdf/10.1142/5566} {https://www.worldscientific.com/doi/pdf/10.1142/5566} \BibitemShut {NoStop}%
\bibitem [{\citenamefont {Zakharov}(1968)}]{Zakharov1968}%
  \BibitemOpen
  \bibfield  {author} {\bibinfo {author} {\bibfnamefont {V.~E.}\ \bibnamefont {Zakharov}},\ }\bibfield  {title} {\enquote {\bibinfo {title} {Stability of periodic waves of finite amplitude on the surface of a deep fluid},}\ }\href {https://api.semanticscholar.org/CorpusID:55755251} {\bibfield  {journal} {\bibinfo  {journal} {Journal of Applied Mechanics and Technical Physics}\ }\textbf {\bibinfo {volume} {9}},\ \bibinfo {pages} {190--194} (\bibinfo {year} {1968})}\BibitemShut {NoStop}%
\bibitem [{\citenamefont {Lünser}\ \emph {et~al.}(2022)\citenamefont {Lünser}, \citenamefont {Hartmann}, \citenamefont {Desmars}, \citenamefont {Behrendt}, \citenamefont {Hoffmann},\ and\ \citenamefont {Klein}}]{Luenser2022}%
  \BibitemOpen
  \bibfield  {author} {\bibinfo {author} {\bibfnamefont {H.}~\bibnamefont {Lünser}}, \bibinfo {author} {\bibfnamefont {M.}~\bibnamefont {Hartmann}}, \bibinfo {author} {\bibfnamefont {N.}~\bibnamefont {Desmars}}, \bibinfo {author} {\bibfnamefont {J.}~\bibnamefont {Behrendt}}, \bibinfo {author} {\bibfnamefont {N.}~\bibnamefont {Hoffmann}}, \ and\ \bibinfo {author} {\bibfnamefont {M.}~\bibnamefont {Klein}},\ }\bibfield  {title} {\enquote {\bibinfo {title} {The influence of characteristic sea state parameters on the accuracy of irregular wave field simulations of different complexity},}\ }\href {\doibase 10.3390/fluids7070243} {\bibfield  {journal} {\bibinfo  {journal} {Fluids}\ }\textbf {\bibinfo {volume} {7}} (\bibinfo {year} {2022}),\ 10.3390/fluids7070243}\BibitemShut {NoStop}%
\bibitem [{\citenamefont {Neill}\ and\ \citenamefont {Hashemi}(2018)}]{Neill2018}%
  \BibitemOpen
  \bibfield  {author} {\bibinfo {author} {\bibfnamefont {S.~P.}\ \bibnamefont {Neill}}\ and\ \bibinfo {author} {\bibfnamefont {M.~R.}\ \bibnamefont {Hashemi}},\ }\bibfield  {title} {\enquote {\bibinfo {title} {Chapter 7 - in situ and remote methods for resource characterization},}\ }in\ \href {\doibase https://doi.org/10.1016/B978-0-12-810448-4.00007-0} {\emph {\bibinfo {booktitle} {Fundamentals of Ocean Renewable Energy}}},\ \bibinfo {series and number} {E-Business Solutions},\ \bibinfo {editor} {edited by\ \bibinfo {editor} {\bibfnamefont {S.~P.}\ \bibnamefont {Neill}}\ and\ \bibinfo {editor} {\bibfnamefont {M.~R.}\ \bibnamefont {Hashemi}}}\ (\bibinfo  {publisher} {Academic Press},\ \bibinfo {year} {2018})\ pp.\ \bibinfo {pages} {157--191}\BibitemShut {NoStop}%
\bibitem [{\citenamefont {{Valenzuela}}(1978)}]{Valenzuela1978}%
  \BibitemOpen
  \bibfield  {author} {\bibinfo {author} {\bibfnamefont {G.~R.}\ \bibnamefont {{Valenzuela}}},\ }\bibfield  {title} {\enquote {\bibinfo {title} {{Theories for the interaction of electromagnetic and oceanic waves {\textemdash} A review}},}\ }\href {\doibase 10.1007/BF00913863} {\bibfield  {journal} {\bibinfo  {journal} {Boundary-Layer Meteorology}\ }\textbf {\bibinfo {volume} {13}},\ \bibinfo {pages} {61--85} (\bibinfo {year} {1978})}\BibitemShut {NoStop}%
\bibitem [{\citenamefont {Dankert}\ and\ \citenamefont {Rosenthal}(2004)}]{Dankert2004}%
  \BibitemOpen
  \bibfield  {author} {\bibinfo {author} {\bibfnamefont {H.}~\bibnamefont {Dankert}}\ and\ \bibinfo {author} {\bibfnamefont {W.}~\bibnamefont {Rosenthal}},\ }\bibfield  {title} {\enquote {\bibinfo {title} {Ocean surface determination from x-band radar-image sequences},}\ }\href {\doibase 10.1029/2003JC002130} {\bibfield  {journal} {\bibinfo  {journal} {Journal of Geophysical Research: Oceans}\ }\textbf {\bibinfo {volume} {109}} (\bibinfo {year} {2004}),\ 10.1029/2003JC002130}\BibitemShut {NoStop}%
\bibitem [{\citenamefont {Borge}\ \emph {et~al.}(2004)\citenamefont {Borge}, \citenamefont {RodrÍguez}, \citenamefont {Hessner},\ and\ \citenamefont {González}}]{NietoBorge2004}%
  \BibitemOpen
  \bibfield  {author} {\bibinfo {author} {\bibfnamefont {J.~N.}\ \bibnamefont {Borge}}, \bibinfo {author} {\bibfnamefont {G.~R.}\ \bibnamefont {RodrÍguez}}, \bibinfo {author} {\bibfnamefont {K.}~\bibnamefont {Hessner}}, \ and\ \bibinfo {author} {\bibfnamefont {P.~I.}\ \bibnamefont {González}},\ }\bibfield  {title} {\enquote {\bibinfo {title} {Inversion of marine radar images for surface wave analysis},}\ }\href {\doibase 10.1175/1520-0426(2004)021<1291:IOMRIF>2.0.CO;2} {\bibfield  {journal} {\bibinfo  {journal} {Journal of Atmospheric and Oceanic Technology}\ }\textbf {\bibinfo {volume} {21}},\ \bibinfo {pages} {1291 -- 1300} (\bibinfo {year} {2004})}\BibitemShut {NoStop}%
\bibitem [{\citenamefont {Salcedo-Sanz}\ \emph {et~al.}(2015)\citenamefont {Salcedo-Sanz}, \citenamefont {{Nieto Borge}}, \citenamefont {Carro-Calvo}, \citenamefont {Cuadra}, \citenamefont {Hessner},\ and\ \citenamefont {Alexandre}}]{Salcedo-Sanz2015}%
  \BibitemOpen
  \bibfield  {author} {\bibinfo {author} {\bibfnamefont {S.}~\bibnamefont {Salcedo-Sanz}}, \bibinfo {author} {\bibfnamefont {J.}~\bibnamefont {{Nieto Borge}}}, \bibinfo {author} {\bibfnamefont {L.}~\bibnamefont {Carro-Calvo}}, \bibinfo {author} {\bibfnamefont {L.}~\bibnamefont {Cuadra}}, \bibinfo {author} {\bibfnamefont {K.}~\bibnamefont {Hessner}}, \ and\ \bibinfo {author} {\bibfnamefont {E.}~\bibnamefont {Alexandre}},\ }\bibfield  {title} {\enquote {\bibinfo {title} {Significant wave height estimation using svr algorithms and shadowing information from simulated and real measured x-band radar images of the sea surface},}\ }\href {\doibase 10.1016/j.oceaneng.2015.04.041} {\bibfield  {journal} {\bibinfo  {journal} {Ocean Engineering}\ }\textbf {\bibinfo {volume} {101}},\ \bibinfo {pages} {244--253} (\bibinfo {year} {2015})}\BibitemShut {NoStop}%
\bibitem [{\citenamefont {Hasselmann}\ \emph {et~al.}(1973)\citenamefont {Hasselmann}, \citenamefont {Barnett}, \citenamefont {Bouws}, \citenamefont {Carlson}, \citenamefont {Cartwright}, \citenamefont {Enke}, \citenamefont {Ewing}, \citenamefont {Gienapp}, \citenamefont {Hasselmann}, \citenamefont {Kruseman}, \citenamefont {Meerburg}, \citenamefont {M{\"u}ller}, \citenamefont {Olbers}, \citenamefont {Richter}, \citenamefont {Sell},\ and\ \citenamefont {Walden.}}]{Hasselmann1973}%
  \BibitemOpen
  \bibfield  {author} {\bibinfo {author} {\bibfnamefont {K.}~\bibnamefont {Hasselmann}}, \bibinfo {author} {\bibfnamefont {T.}~\bibnamefont {Barnett}}, \bibinfo {author} {\bibfnamefont {E.}~\bibnamefont {Bouws}}, \bibinfo {author} {\bibfnamefont {H.}~\bibnamefont {Carlson}}, \bibinfo {author} {\bibfnamefont {D.}~\bibnamefont {Cartwright}}, \bibinfo {author} {\bibfnamefont {K.}~\bibnamefont {Enke}}, \bibinfo {author} {\bibfnamefont {J.}~\bibnamefont {Ewing}}, \bibinfo {author} {\bibfnamefont {H.}~\bibnamefont {Gienapp}}, \bibinfo {author} {\bibfnamefont {D.}~\bibnamefont {Hasselmann}}, \bibinfo {author} {\bibfnamefont {P.}~\bibnamefont {Kruseman}}, \bibinfo {author} {\bibfnamefont {A.}~\bibnamefont {Meerburg}}, \bibinfo {author} {\bibfnamefont {P.}~\bibnamefont {M{\"u}ller}}, \bibinfo {author} {\bibfnamefont {D.}~\bibnamefont {Olbers}}, \bibinfo {author} {\bibfnamefont {K.}~\bibnamefont {Richter}}, \bibinfo {author} {\bibfnamefont {W.}~\bibnamefont {Sell}}, \ and\ \bibinfo {author} {\bibfnamefont
  {H.}~\bibnamefont {Walden.}},\ }\bibfield  {title} {\enquote {\bibinfo {title} {Measurements of wind-wave growth and swell decay during the joint north sea wave project (jonswap)},}\ }\href@noop {} {\bibfield  {journal} {\bibinfo  {journal} {Erg{\"a}nzung zur Deutschen Hydrographischen Zeitschrift, Reihe A (8)}\ }\textbf {\bibinfo {volume} {12}},\ \bibinfo {pages} {1--95} (\bibinfo {year} {1973})}\BibitemShut {NoStop}%
\bibitem [{\citenamefont {Bouws}\ \emph {et~al.}(1985)\citenamefont {Bouws}, \citenamefont {G{\"u}nther}, \citenamefont {Rosenthal},\ and\ \citenamefont {Vincent}}]{Bouws1985}%
  \BibitemOpen
  \bibfield  {author} {\bibinfo {author} {\bibfnamefont {E.}~\bibnamefont {Bouws}}, \bibinfo {author} {\bibfnamefont {H.}~\bibnamefont {G{\"u}nther}}, \bibinfo {author} {\bibfnamefont {W.}~\bibnamefont {Rosenthal}}, \ and\ \bibinfo {author} {\bibfnamefont {C.~L.}\ \bibnamefont {Vincent}},\ }\bibfield  {title} {\enquote {\bibinfo {title} {Similarity of the wind wave spectrum in finite depth water: 1. spectral form},}\ }\href {\doibase 10.1029/JC090iC01p00975} {\bibfield  {journal} {\bibinfo  {journal} {Journal of Geophysical Research}\ }\textbf {\bibinfo {volume} {90}},\ \bibinfo {pages} {975--986} (\bibinfo {year} {1985})}\BibitemShut {NoStop}%
\bibitem [{\citenamefont {Chen}\ \emph {et~al.}(2018)\citenamefont {Chen}, \citenamefont {Badrinarayanan}, \citenamefont {Lee},\ and\ \citenamefont {Rabinovich}}]{Chen2018}%
  \BibitemOpen
  \bibfield  {author} {\bibinfo {author} {\bibfnamefont {Z.}~\bibnamefont {Chen}}, \bibinfo {author} {\bibfnamefont {V.}~\bibnamefont {Badrinarayanan}}, \bibinfo {author} {\bibfnamefont {C.-Y.}\ \bibnamefont {Lee}}, \ and\ \bibinfo {author} {\bibfnamefont {A.}~\bibnamefont {Rabinovich}},\ }\href@noop {} {\enquote {\bibinfo {title} {Gradnorm: Gradient normalization for adaptive loss balancing in deep multitask networks},}\ } (\bibinfo {year} {2018}),\ \Eprint {http://arxiv.org/abs/1711.02257} {arXiv:1711.02257 [cs.CV]} \BibitemShut {NoStop}%
\bibitem [{\citenamefont {Heydari}, \citenamefont {Thompson},\ and\ \citenamefont {Mehmood}(2019)}]{Heydari2019}%
  \BibitemOpen
  \bibfield  {author} {\bibinfo {author} {\bibfnamefont {A.~A.}\ \bibnamefont {Heydari}}, \bibinfo {author} {\bibfnamefont {C.~A.}\ \bibnamefont {Thompson}}, \ and\ \bibinfo {author} {\bibfnamefont {A.}~\bibnamefont {Mehmood}},\ }\href@noop {} {\enquote {\bibinfo {title} {Softadapt: Techniques for adaptive loss weighting of neural networks with multi-part loss functions},}\ } (\bibinfo {year} {2019}),\ \Eprint {http://arxiv.org/abs/1912.12355} {arXiv:1912.12355 [cs.LG]} \BibitemShut {NoStop}%
\bibitem [{\citenamefont {Wang}, \citenamefont {Teng},\ and\ \citenamefont {Perdikaris}(2020)}]{Wang2020}%
  \BibitemOpen
  \bibfield  {author} {\bibinfo {author} {\bibfnamefont {S.}~\bibnamefont {Wang}}, \bibinfo {author} {\bibfnamefont {Y.}~\bibnamefont {Teng}}, \ and\ \bibinfo {author} {\bibfnamefont {P.}~\bibnamefont {Perdikaris}},\ }\href {https://arxiv.org/abs/2001.04536} {\enquote {\bibinfo {title} {Understanding and mitigating gradient pathologies in physics-informed neural networks},}\ } (\bibinfo {year} {2020}),\ \Eprint {http://arxiv.org/abs/2001.04536} {2001.04536} \BibitemShut {NoStop}%
\bibitem [{\citenamefont {Bischof}\ and\ \citenamefont {Kraus}(2021)}]{Bischof2021}%
  \BibitemOpen
  \bibfield  {author} {\bibinfo {author} {\bibfnamefont {R.}~\bibnamefont {Bischof}}\ and\ \bibinfo {author} {\bibfnamefont {M.}~\bibnamefont {Kraus}},\ }\bibfield  {title} {\enquote {\bibinfo {title} {Multi-objective loss balancing for physics-informed deep learning},}\ }\href {https://api.semanticscholar.org/CorpusID:239024424} {\bibfield  {journal} {\bibinfo  {journal} {ArXiv}\ }\textbf {\bibinfo {volume} {abs/2110.09813}} (\bibinfo {year} {2021})}\BibitemShut {NoStop}%
\bibitem [{\citenamefont {Kingma}\ and\ \citenamefont {Ba}(2017)}]{Kingma2017}%
  \BibitemOpen
  \bibfield  {author} {\bibinfo {author} {\bibfnamefont {D.~P.}\ \bibnamefont {Kingma}}\ and\ \bibinfo {author} {\bibfnamefont {J.}~\bibnamefont {Ba}},\ }\href {https://arxiv.org/abs/1412.6980} {\enquote {\bibinfo {title} {Adam: A method for stochastic optimization},}\ } (\bibinfo {year} {2017}),\ \Eprint {http://arxiv.org/abs/1412.6980} {arXiv:1412.6980 [cs.LG]} \BibitemShut {NoStop}%
\bibitem [{\citenamefont {Loshchilov}\ and\ \citenamefont {Hutter}(2019)}]{Loshchilov2019}%
  \BibitemOpen
  \bibfield  {author} {\bibinfo {author} {\bibfnamefont {I.}~\bibnamefont {Loshchilov}}\ and\ \bibinfo {author} {\bibfnamefont {F.}~\bibnamefont {Hutter}},\ }\href {https://arxiv.org/abs/1711.05101} {\enquote {\bibinfo {title} {Decoupled weight decay regularization},}\ } (\bibinfo {year} {2019}),\ \Eprint {http://arxiv.org/abs/1711.05101} {arXiv:1711.05101 [cs.LG]} \BibitemShut {NoStop}%
\bibitem [{\citenamefont {Perlin}\ and\ \citenamefont {Bustamante}(2016)}]{Perlin2014}%
  \BibitemOpen
  \bibfield  {author} {\bibinfo {author} {\bibfnamefont {M.}~\bibnamefont {Perlin}}\ and\ \bibinfo {author} {\bibfnamefont {M.}~\bibnamefont {Bustamante}},\ }\bibfield  {title} {\enquote {\bibinfo {title} {A robust quantitative comparison criterion of two signals based on the sobolev norm of their difference},}\ }\href {\doibase 10.1007/s10665-016-9849-7} {\bibfield  {journal} {\bibinfo  {journal} {Journal of Engineering Mathematics}\ }\textbf {\bibinfo {volume} {101}},\ \bibinfo {pages} {115--124} (\bibinfo {year} {2016})}\BibitemShut {NoStop}%
\bibitem [{\citenamefont {Kim}\ \emph {et~al.}(2023)\citenamefont {Kim}, \citenamefont {Ducrozet}, \citenamefont {Bonnefoy}, \citenamefont {Leroy},\ and\ \citenamefont {Perignon}}]{Kim2023}%
  \BibitemOpen
  \bibfield  {author} {\bibinfo {author} {\bibfnamefont {I.-C.}\ \bibnamefont {Kim}}, \bibinfo {author} {\bibfnamefont {G.}~\bibnamefont {Ducrozet}}, \bibinfo {author} {\bibfnamefont {F.}~\bibnamefont {Bonnefoy}}, \bibinfo {author} {\bibfnamefont {V.}~\bibnamefont {Leroy}}, \ and\ \bibinfo {author} {\bibfnamefont {Y.}~\bibnamefont {Perignon}},\ }\bibfield  {title} {\enquote {\bibinfo {title} {Real-time phase-resolved ocean wave prediction in directional wave fields: Enhanced algorithm and experimental validation},}\ }\href {\doibase 10.1016/j.oceaneng.2023.114212} {\bibfield  {journal} {\bibinfo  {journal} {Ocean Engineering}\ }\textbf {\bibinfo {volume} {276}},\ \bibinfo {pages} {114212} (\bibinfo {year} {2023})}\BibitemShut {NoStop}%
\bibitem [{\citenamefont {Dermatis}\ \emph {et~al.}(2025)\citenamefont {Dermatis}, \citenamefont {Lasbleis}, \citenamefont {Kim}, \citenamefont {{De Hauteclocque}}, \citenamefont {Bouscasse},\ and\ \citenamefont {Ducrozet}}]{Dermatis2025}%
  \BibitemOpen
  \bibfield  {author} {\bibinfo {author} {\bibfnamefont {A.}~\bibnamefont {Dermatis}}, \bibinfo {author} {\bibfnamefont {M.}~\bibnamefont {Lasbleis}}, \bibinfo {author} {\bibfnamefont {S.}~\bibnamefont {Kim}}, \bibinfo {author} {\bibfnamefont {G.}~\bibnamefont {{De Hauteclocque}}}, \bibinfo {author} {\bibfnamefont {B.}~\bibnamefont {Bouscasse}}, \ and\ \bibinfo {author} {\bibfnamefont {G.}~\bibnamefont {Ducrozet}},\ }\bibfield  {title} {\enquote {\bibinfo {title} {A multi-fidelity approach for the evaluation of extreme wave loads using nonlinear response-conditioned waves},}\ }\href {\doibase https:/doi.org/10.1016/j.oceaneng.2024.119919} {\bibfield  {journal} {\bibinfo  {journal} {Ocean Engineering}\ }\textbf {\bibinfo {volume} {316}},\ \bibinfo {pages} {119919} (\bibinfo {year} {2025})}\BibitemShut {NoStop}%
\bibitem [{\citenamefont {van Essen}(2021)}]{vanEssen2021}%
  \BibitemOpen
  \bibfield  {author} {\bibinfo {author} {\bibfnamefont {S.}~\bibnamefont {van Essen}},\ }\bibfield  {title} {\enquote {\bibinfo {title} {Influence of wave variability on ship response during deterministically repeated seakeeping tests at forward speed},}\ }in\ \href@noop {} {\emph {\bibinfo {booktitle} {Practical Design of Ships and Other Floating Structures}}},\ \bibinfo {editor} {edited by\ \bibinfo {editor} {\bibfnamefont {T.}~\bibnamefont {Okada}}, \bibinfo {editor} {\bibfnamefont {K.}~\bibnamefont {Suzuki}}, \ and\ \bibinfo {editor} {\bibfnamefont {Y.}~\bibnamefont {Kawamura}}}\ (\bibinfo  {publisher} {Springer Singapore},\ \bibinfo {address} {Singapore},\ \bibinfo {year} {2021})\ pp.\ \bibinfo {pages} {899--925}\BibitemShut {NoStop}%
\end{thebibliography}%

\end{document}